%% file: main.tex
\documentclass[runningheads]{llncs}

\usepackage{eccv}

\usepackage{eccvabbrv}

\usepackage{graphicx}   
\usepackage{amsmath}
\usepackage{amssymb}
\usepackage{booktabs}
\usepackage{multirow}
\usepackage{colortbl}
\usepackage{arydshln} 
\usepackage{makecell}
\usepackage{listings}
\usepackage{subcaption}
\usepackage{pifont}
\usepackage{adjustbox}
\usepackage{array}
\usepackage[accsupp]{axessibility}  

\usepackage{hyperref}

\usepackage{orcidlink}

\begin{document}

\title{TRACE: Task-Adaptive Reasoning and Representation Learning for Universal Multimodal Retrieval} 

\titlerunning{Abbreviated paper title}

\author{Xiangzhao Hao\inst{1,2}\thanks{These authors contributed equally to this work.} \and
Shijie Wang\inst{1,2}\protect\footnotemark[1] \and 
Tianyu Yang\inst{1,2}\protect\footnotemark[1] \and
Tianyue Wang\inst{1,2} \and
Haiyun Guo\inst{1,2} \and
Jinqiao Wang\inst{1,2}}

\authorrunning{X.~Hao et al.}

\institute{Institute of Automation, Chinese Academy of Sciences \and
University of Chinese Academy of Sciences\\
\email{\{haoxiangzhao2023, wangshijie2026, yangtianyu2024\}@ia.ac.cn, wangtianyue25@mails.ucas.ac.cn, \{haiyun.guo, jqwang\}@nlpr.ia.ac.cn}}
\maketitle

\begin{abstract}
Universal Multimodal Retrieval requires unified embedding models capable of interpreting diverse user intents, ranging from simple keywords to complex compositional instructions. While Multimodal Large Language Models (MLLMs) possess strong reasoning capabilities, prevailing adaptations confine them to static encoders, underutilizing their generative potential. This encoder-only paradigm struggles with complex intents that demand logical deduction rather than superficial pattern matching. To address this, we introduce \textbf{TRACE} (\textbf{T}ask-adaptive \textbf{R}easoning \textbf{A}nd \textbf{C}ompressing \textbf{E}mbeddings). TRACE unifies generative reasoning with discriminative representation learning. It first generates a structured Chain-of-Thought (CoT) to explicitly reason about the query, and subsequently compresses this reasoning trace into a compact embedding via a dedicated token. To train this framework, we construct \textbf{M-BEIR-CoT}, a large-scale dataset featuring a difficulty-aware routing strategy. Experiments on the M-BEIR benchmark establish TRACE as the new state-of-the-art. Crucially, TRACE demonstrates a learned implicit routing behavior. It autonomously activates reasoning for complex queries while bypassing it for simpler ones, achieving an optimal balance between retrieval accuracy and inference throughput. Furthermore, by internalizing the deductive process, TRACE exhibits remarkable zero-shot transferability to unseen domains and novel constraints.

\end{abstract}

\begin{figure}[tb]
  \centering
    \includegraphics[width=\textwidth]{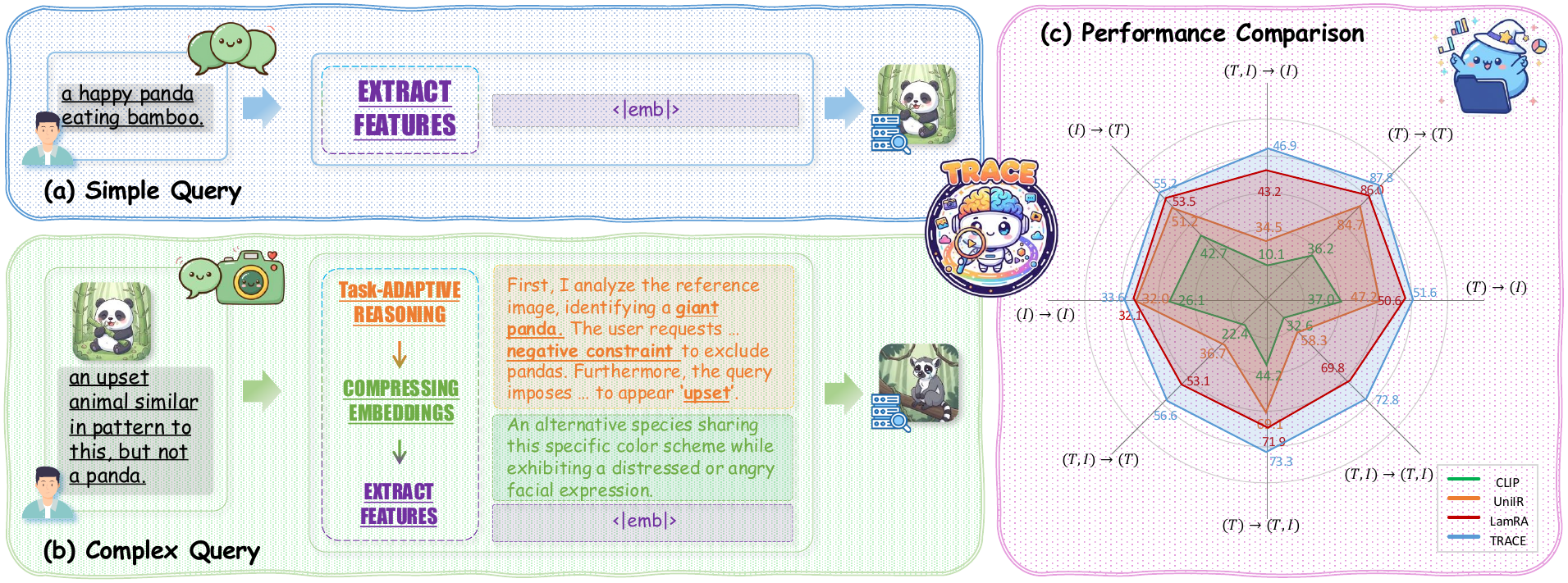}
\caption{
\textbf{The TRACE Framework.} TRACE learns a query-dependent inference strategy. 
\textbf{(a)} For simple queries, it implicitly bypasses the reasoning stage and directly extracts features to maintain high efficiency. 
\textbf{(b)} For complex queries, it automatically activates the task-adaptive reasoning process. The model generates an explicit reasoning trace~\cite{cot} to resolve semantic ambiguities before compressing this context into the final representation. 
\textbf{(c)} Performance comparison on the M-BEIR benchmark~\cite{wei2023uniir} demonstrates the effectiveness of TRACE, particularly on reasoning-intensive tasks.
}
    \label{fig:teaser}
\end{figure}

\section{Introduction}

\textbf{The Bottleneck of Current Universal Retrieval.} Universal Multimodal Retrieval aims to unify search across diverse modalities, ranging from pure text to complex and interleaved image-text queries~\cite{wei2023uniir, lin2024mm, jiang2024vlm2vec, zhang2024magiclens, qwen2vl,yangtianyu}. Recently, Multimodal Large Language Models have revolutionized cross-modal understanding~\cite{li2023blip, liu2023visual, openai2023gpt4v}. However, the dominant paradigm for adapting these models to retrieval tasks treats them as static encoders~\cite{wei2023uniir, jiang2024e5, lin2024mm, jiang2024vlm2vec, zhang2024gme}. The model ingests multimodal inputs and compresses them directly into a fixed-dimensional embedding via a single forward pass. While efficient for simple matching~\cite{radford2021learning, jia2021scaling, yu2022coca, zhai2023sigmoid, flip}, this approach faces a critical bottleneck when handling compositional user intents~\cite{vo2019composing, wu2021fashion, anwaar2021compositional, baldrati2023zero, vaze2023genecis}, such as commands to remove specific objects or alter visual attributes. Forcing a model to implicitly perform multi-step logic within a single encoding step creates a severe \textit{cognitive bottleneck}, fundamentally underutilizing its inherent generative reasoning capacity.

\textbf{A Paradigm Shift: Reasoning then Encoding.} To unlock the full potential of these models in retrieval, we propose a shift toward task-adaptive reasoning prior to feature representation. Inspired by the success of structured reasoning in natural language processing~\cite{cot, xu2024llava}, we argue that an explicit reasoning step is essential for complex scenarios. Instead of directly encoding the query, the model should first utilize its generative head to produce a reasoning trace that articulates its understanding of the user intent~\cite{query_rewriting, ye2023enhancingconversationalsearchlarge}. This generated context acts as a semantic bridge, guiding the model to produce a high-fidelity embedding that captures the nuanced reasoning state rather than surface-level features alone~\cite{bert, gao2021simcse}.

\textbf{The TRACE Framework.} Building on this insight, we introduce TRACE. This novel framework seamlessly integrates generative reasoning with discriminative representation learning. TRACE first leverages the language model to generate a structured reasoning path that analyzes and fuses query modalities~\cite{xu2024llava, liu2025lamra, sun2025leveraging, Tang2024OR}. Subsequently, it employs a dedicated token and causal attention mechanisms to compress this explicit trace into a compact vector. Crucially, TRACE learns an adaptive routing mechanism. It automatically discerns query difficulty, bypassing the reasoning stage for simple keyword searches while activating the reasoning process for complex tasks~\cite{miech2021thinking}. This ensures an optimal balance between performance and inference efficiency without requiring explicit architectural branching.

\textbf{Data Construction and Core Findings.} A major hurdle in training such a system is the scarcity of datasets containing high-quality reasoning traces aligned with retrieval targets. To address this, we construct M-BEIR-CoT, a large-scale dataset synthesized via foundation models~\cite{megapairs, chen2023sharegpt4v} based on the M-BEIR benchmark~\cite{wei2023uniir}. A rigorous filtration process ensures the reasoning is supportive rather than hallucinatory. 

To summarize, our main contributions are:
\begin{itemize}
    \item We propose TRACE, a universal retrieval framework that explicitly integrates task-adaptive reasoning into the discriminative embedding process. Unlike traditional two-stage pipelines, TRACE internalizes reasoning to balance accuracy and inference throughput.
    \item We introduce M-BEIR-CoT, a large-scale and quality-filtered dataset designed to foster adaptive reasoning capabilities for retrieval tasks, addressing the critical data scarcity in this domain.
    \item We establish new state-of-the-art performance on the M-BEIR benchmark and extensive zero-shot scenarios. Furthermore, we uncover a fundamental asymmetry in reasoning for retrieval: while query-side reasoning significantly enhances semantic alignment, forcing candidate-side reasoning catastrophically degrades performance by overfitting to generated text patterns.
\end{itemize}

\section{Related Work}
\label{sec:related}

\subsection{Evolution of Universal Multimodal Retrieval}
The field of multimodal retrieval has progressed from specialized dual-encoder architectures to unified generative frameworks. Early dominant approaches, such as CLIP~\cite{radford2021learning} and ALIGN~\cite{jia2021scaling}, established a strong baseline by aligning visual and textual representations through contrastive learning on massive datasets. While effective for standard image-text matching, these dual-encoder models process modalities independently before a final late interaction. This architecture fundamentally limits their ability to capture fine-grained compositional logic, such as composed image retrieval tasks requiring object modification while preserving background context~\cite{vo2019composing, wu2021fashion, anwaar2021compositional,wangtianyue}.

To address these limitations, recent research has pivoted towards Multimodal Large Language Models (MLLMs) for representation learning. Methods like UniIR~\cite{wei2023uniir}, E5-V~\cite{jiang2024e5}, and LamRA~\cite{liu2025lamra} adapt MLLMs to serve as universal retrievers, typically by appending specific prompts to extract embeddings from the last hidden state~\cite{lin2024mm, jiang2024vlm2vec, zhang2024gme}. Although these methods leverage the broad world knowledge of MLLMs to outperform dual-encoders on zero-shot benchmarks, they predominantly treat the model as a static encoder. By compressing inputs directly into embeddings via a single forward pass, these approaches bypass the inherent generative reasoning capabilities of the backbone. This creates a cognitive bottleneck, where the model is forced to map complex logical queries directly to a vector space without intermediate deductive steps.

\subsection{Chain-of-Thought for Discriminative Tasks}
Chain-of-Thought (CoT) prompting~\cite{cot, xu2024llava, chu-etal-2024-navigate} has proven effective in enhancing the reasoning capabilities of Large Language Models across various generative tasks. In the multimodal domain, recent works have successfully extended CoT to reduce hallucinations and improve interpretability in visual question answering and captioning~\cite{xu2024llava, lai2024lisa, pi2023detgpt}. However, the integration of explicit reasoning into discriminative retrieval remains underexplored.

Existing approaches that incorporate reasoning into retrieval typically rely on disjointed, multi-stage pipelines. For instance, some methods utilize a generative model for external query expansion or rewriting~\cite{query_rewriting, ye2023enhancingconversationalsearchlarge} before feeding the augmented text into a separate encoder. This separation prevents the seamless flow of visual perception and logical deduction. TRACE distinguishes itself by internalizing this process within a unified end-to-end framework. Rather than treating reasoning and encoding as separate steps handled by different models, TRACE generates a latent reasoning trajectory that is directly compressed into the retrieval embedding. This design allows the model to adaptively balance inference efficiency with the depth of reasoning required for complex queries.

\section{Method}
\label{sec:method}
In this section, we detail the proposed TRACE framework. We begin by formulating the problem of reasoning-aware universal retrieval in \cref{sec:formulation}. \Cref{sec:data_construction} introduces the construction of our M-BEIR-CoT dataset, which serves as the cognitive foundation for our model. Finally, in \cref{sec:framework}, we present the unified TRACE architecture and its single-stage training strategy.

\begin{figure*}[t]
  \centering
  \includegraphics[width=\textwidth]{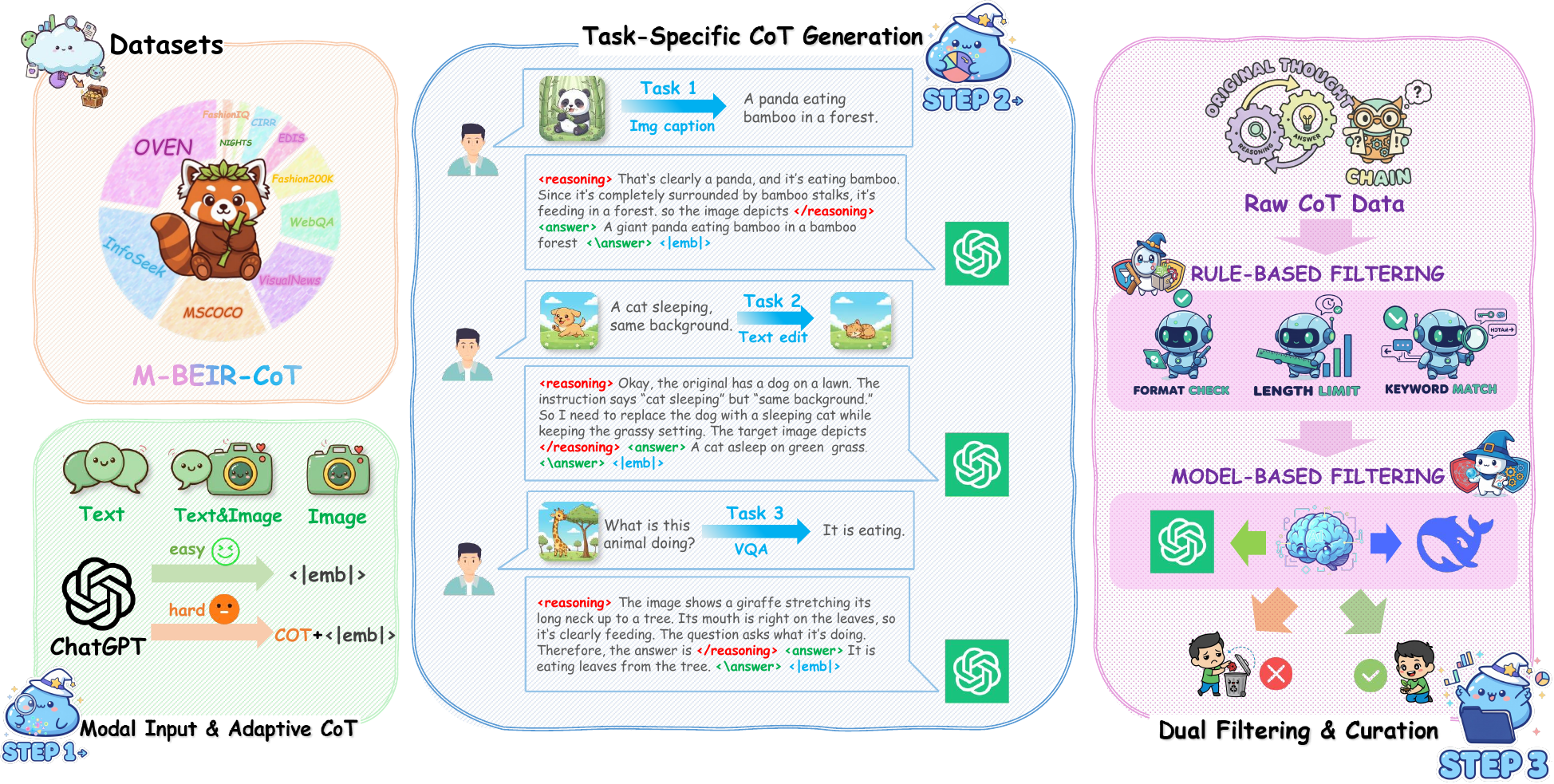}
  \caption{\textbf{The construction pipeline of the M-BEIR-CoT dataset.} The process operates in three phases: 
  \textbf{(1) Query Complexity Assessment:} An advanced MLLM assesses query difficulty, routing simple queries to a direct path (generating only \texttt{<|emb|>}) and complex queries to a reasoning path (generating \texttt{CoT} + \texttt{<|emb|>}).
  \textbf{(2) Task-Specific CoT Generation:} We design specialized prompts for diverse tasks (\eg, captioning, text edit, VQA) to generate structured reasoning traces enclosed in \texttt{<reasoning>} tags.
  \textbf{(3) Dual Filtering \& Curation:} To ensure data quality, we apply a coarse-to-fine strategy. We first use rule-based filtering to verify formats and lengths, followed by model-based filtering to ensure semantic consistency between the generated text and ground-truth targets.}
  \label{fig:data_pipeline}
\end{figure*}

\subsection{Problem Formulation}
\label{sec:formulation}
We aim to learn a universal retrieval function $f_\theta$ capable of handling multimodal queries $Q$ (comprising text $T$, images $I$, or interleaved sequences) to retrieve a target $C$ from a candidate set $\Omega$. Unlike traditional retrievers that map inputs directly to a static vector ($Q \to \mathbf{v}_q$), TRACE treats retrieval as a conditional generation-then-compression process. 

Formally, given a query $Q$, the model generates an intermediate sequence $S$ ending with a special token \texttt{<|emb|>}:
\begin{equation}
    S = [\mathcal{R} ; \texttt{<|emb|>}], \quad \text{where } \mathcal{R} = 
    \begin{cases} 
        \emptyset & \text{if } z=0 \\ 
        \{r_1, \dots, r_k\} & \text{if } z=1 
    \end{cases}
\end{equation}
Here, $z \in \{0, 1\}$ represents a latent complexity variable determining whether to activate the reasoning trace $\mathcal{R}$. The final query representation $\mathbf{e}_q$ is extracted from the hidden state responsible for predicting the \texttt{<|emb|>} token. The goal is to maximize the similarity score $\text{sim}(\mathbf{e}_q, \mathbf{e}_c)$ with the ground-truth target.

\subsection{Construction of M-BEIR-CoT}
\label{sec:data_construction}
Existing retrieval datasets typically lack the explicit logic chains required to train a reasoning-aware model. To bridge this gap, we construct M-BEIR-CoT, a large-scale dataset built upon the M-BEIR benchmark~\cite{wei2023uniir}. Our construction pipeline proceeds in three distinct phases, as illustrated in Fig.~\ref{fig:data_pipeline}.

\textbf{Phase 1: Query Complexity Assessment \& Adaptive Routing.} To optimize the trade-off between efficiency and reasoning depth, we employ an advanced MLLM (\eg, GPT-4o~\cite{openai2023gpt4v}) as a query complexity assessor. This module categorizes modal inputs into two streams: a direct encoding stream for simple pattern-matching tasks ($z=0$), and a reasoning-augmented stream for complex queries involving constraints or logic ($z=1$). This creates a hybrid dataset structure that teaches the model when to reason and when to act reflexively.

\textbf{Phase 2: Task-Specific CoT Generation.} For queries routed to the reasoning stream, we employ specialized prompt templates tailored to different retrieval sub-tasks. For visual reasoning, the model articulates fine-grained details before summarization. For instruction following (\eg, CIR~\cite{vo2019composing, wu2021fashion}), the reasoning trace explicitly defines the source state, the target operation, and the invariant context. For logical deduction, the model performs multi-hop inference. All outputs are strictly formatted with \texttt{<reasoning>} and \texttt{<answer>} tags to facilitate structured parsing.

\textbf{Phase 3: Dual Filtering \& Curation.} To mitigate hallucinations, we implement a rigorous coarse-to-fine filtration protocol. We first apply rule-based filtering to remove samples with invalid formatting or insufficient length. Subsequently, a model-based consistency check is performed using a strong verifier model to compute the semantic alignment between the generated answer and the ground-truth target. Only samples surpassing a high confidence threshold are retained, ensuring the dataset provides high-quality supervision for both generation and retrieval.

Following this pipeline, we curated 575,442 high-quality reasoning samples. Crucially, we remove the auxiliary tags (\texttt{<reasoning>}, \texttt{<answer>}) during training to enforce natural generation. Merging these with 518,311 simple samples yields our final M-BEIR-CoT dataset. Further details regarding the dataset construction are provided in the supplementary material.


\begin{figure*}[t]
  \centering
  \includegraphics[width=0.8\textwidth]{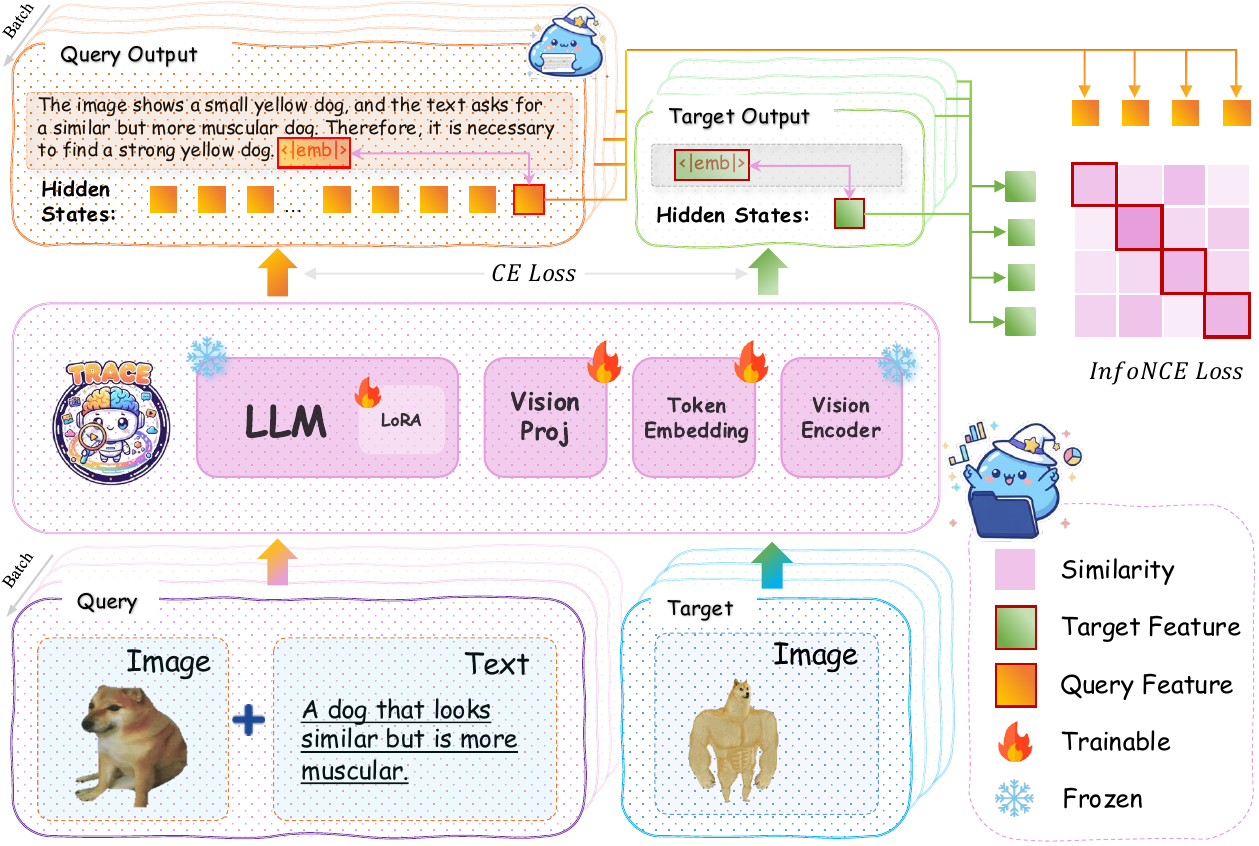} 
  \caption{\textbf{Illustration of the TRACE architecture.} 
  The model processes a multimodal query through a frozen vision encoder and a trainable projector. 
  The LLM acts as a unified reasoner and encoder. 
  It first generates a Chain-of-Thought (CoT)~\cite{cot} to interpret the intent and then compresses the semantics into a learnable \texttt{<|emb|>} token. 
  The final query feature is extracted from the hidden state immediately preceding \texttt{<|emb|>}. 
  During training, the model is optimized jointly using Cross-Entropy (CE) loss for reasoning generation and InfoNCE loss~\cite{oord2018representation} for embedding alignment.}
  \label{fig:method_arch}
\end{figure*}

\subsection{The TRACE Framework}
\label{sec:framework}

\textbf{Architecture and Adaptive Mechanism.}
Our framework is built upon Qwen2.5-VL~\cite{qwen2.5vl}, consisting of a vision encoder, a projector, and a Large Language Model (LLM) backbone. TRACE processes an input query $Q$ through the LLM to autoregressively generate a response sequence. 

A unique feature of our architecture is the emergent adaptive activation of reasoning. Because the M-BEIR-CoT dataset contains a curated mix of direct encodings ($z=0$) and reasoning-augmented sequences ($z=1$), the model implicitly learns to evaluate query complexity. During inference, the model dynamically determines the optimal path via standard autoregressive decoding. Let $\mathcal{V}_{\text{text}}$ be the standard text vocabulary. At the first decoding step, the output sequence is governed by:
\small
\begin{equation}
    \text{Output}(Q) = 
    \begin{cases} 
        [\texttt{<|emb|>}] & \text{if } \texttt{<|emb|>} = \arg\max_{y \in \mathcal{V}} P(y \mid Q) \\
        [\text{CoT Tokens}, \texttt{<|emb|>}] & \text{if } \arg\max_{y \in \mathcal{V}} P(y \mid Q) \in \mathcal{V}_{\text{text}}
    \end{cases}
\end{equation}
\normalsize
Instead of relying on explicit architectural branching or a manually tuned gating network, TRACE naturally shifts its initial probability mass to the \texttt{<|emb|>} token for simple queries, and to text tokens for complex ones.

Due to the causal attention mechanism, the hidden state $\mathbf{h}_t$ is optimized to predict the next token $y_{t+1}$. Therefore, the token immediately preceding \texttt{<|emb|>} is responsible for predicting this end-of-sequence identifier. Consequently, this specific state aggregates information from the entire preceding context encompassing both the raw query and the generated CoT—acting as the optimal semantic bottleneck. We extract the hidden state of this pre-token as the final retrieval embedding $\mathbf{e}_q \in \mathbb{R}^d$.

\textbf{Unified Single-Stage Training.}
We employ a unified single-stage training strategy that optimizes the model on the hybrid M-BEIR-CoT dataset. This approach equips the model with both reasoning and representation capabilities simultaneously using a hybrid objective function.

First, the \textbf{Generative Reasoning Loss ($\mathcal{L}_{\text{gen}}$)} utilizes standard cross-entropy loss to supervise the generation of reasoning trace tokens, forcing the model to internalize the logic of intent decomposition:
\begin{equation}
    \mathcal{L}_{\text{gen}} = - \sum_{t=1}^{|\mathcal{R}|} \log P(y_t \mid y_{<t}, Q),
\end{equation}
where $y$ represents the ground-truth reasoning tokens.

Second, the \textbf{Discriminative Contrastive Loss ($\mathcal{L}_{\text{ret}}$)} structures the embedding space using InfoNCE loss~\cite{oord2018representation} on the final \texttt{<|emb|>} token. Given a batch of $B$ query-target pairs $\{(Q_i, C_i)\}_{i=1}^B$:
\begin{equation}
    \mathcal{L}_{\text{ret}} = - \frac{1}{B} \sum_{i=1}^{B} \log \frac{\exp(\text{sim}(\mathbf{e}_{q_i}, \mathbf{e}_{c_i}) / \tau)}{\sum_{j=1}^{B} \exp(\text{sim}(\mathbf{e}_{q_i}, \mathbf{e}_{c_j}) / \tau)}.
\end{equation}
The final objective is a weighted sum: 
\begin{equation}
    \mathcal{L} = \lambda_{\text{gen}} \mathcal{L}_{\text{gen}} + \lambda_{\text{ret}} \mathcal{L}_{\text{ret}}
\end{equation}
This joint optimization ensures that the generative reasoning process is explicitly guided to maximize the discriminative power of the final retrieval embedding.
\section{Experiments}
\label{sec:experiments}

\subsection{Experimental Setup}
\label{sec:experimental_setup}

\textbf{Datasets and Metrics.} 
We primarily utilize our constructed M-BEIR-CoT dataset for instruction tuning. This dataset is derived from the M-BEIR~\cite{wei2023uniir} benchmark, enriched with high-quality reasoning annotations to support cognitive retrieval. For in-domain evaluation, we conduct assessments on the standard M-BEIR test set, which encompasses eight distinct retrieval tasks across 10 different datasets. To rigorously evaluate generalization capabilities, we investigate performance on a suite of 13 unseen datasets strictly excluded from the training phase. These include ShareGPT4V~\cite{chen2023sharegpt4v}, Urban-1K~\cite{zhang2024long}, CIRCO~\cite{baldrati2023zero}, and Visual Dialog~\cite{das2017visual}, covering diverse domains such as fine-grained recognition, composed image retrieval, and multi-turn dialogue. We adhere to standard evaluation metrics. For retrieval tasks, we report Recall@K, and for image-text matching tasks, we employ Accuracy as the primary metric.

\input{tables/mbeir}

\textbf{Baselines.} 
We compare TRACE against a comprehensive set of state-of-the-art methods. The first group comprises general-purpose vision-language models and dual-encoders, including CLIP~\cite{radford2021learning}, SigLIP~\cite{zhai2023sigmoid}, BLIP2~\cite{li2023blip}, and the base Qwen2.5-VL~\cite{qwen2.5vl}. These baselines benchmark fundamental visual-semantic alignment. The second group consists of specialized universal retrievers such as UniIR~\cite{wei2023uniir} and the recent state-of-the-art LamRA~\cite{liu2025lamra}. This comparison validates the effectiveness of our approach within the specific context of unified multimodal search.

\textbf{Implementation Details.} 
Our framework leverages Qwen2.5-VL-7B~\cite{qwen2.5vl} as the primary backbone. Distinct from complex multi-stage pipelines, we employ a unified single-stage training strategy. The model is fine-tuned on the M-BEIR-CoT dataset using 16 NVIDIA H20 GPUs. We train for one epoch with a global batch size of 1024 and a learning rate of $2\times 10^{-4}$, utilizing the AdamW optimizer with a cosine learning rate decay schedule. To handle diverse task types smoothly, we apply a task-aware sampling strategy to ensure balanced gradient updates across different modalities. The parameters of the vision encoder remain frozen to preserve pre-trained visual representations, while the language model parameters are updated using Low-Rank Adaptation~\cite{lora}.

\subsection{Efficiency and Adaptive Analysis}
\label{sec:efficiency}

A primary concern when integrating generative reasoning into retrieval systems is the potential latency overhead. To address this, we conduct a rigorous analysis comparing query throughput and retrieval accuracy across different task difficulties. We select MSCOCO to represent simple pattern-matching queries and CIRR to represent complex reasoning-intensive queries.

\begin{table}[tb]
    \centering
    \caption{\textbf{Efficiency vs. Accuracy Trade-off.} We compare standard direct embedding, forced reasoning (Always CoT), and our adaptive approach (TRACE). QPS (Queries Per Second) is measured on a single NVIDIA H20 GPU during the online encoding phase.}
    \begin{tabular}{lcccc}
    \toprule
    \multirow{2}{*}{Method} & \multicolumn{2}{c}{\textbf{MSCOCO (Simple)}} & \multicolumn{2}{c}{\textbf{CIRR (Complex)}} \\
    \cmidrule(lr){2-3} \cmidrule(lr){4-5}
     & QPS $\uparrow$ & R@5 $\uparrow$ & QPS $\uparrow$ & R@5 $\uparrow$ \\
    \midrule
    Direct Embedding & \textbf{15.68} & 87.40 & \textbf{12.15} & 53.06 \\
    Always CoT (Forced) & 4.45 & 63.90 & 3.42 & 54.73 \\
    \textbf{TRACE (Ours)} & 8.25 & \textbf{89.10} & 6.48 & \textbf{57.03} \\
    \bottomrule
    \end{tabular}
    \label{tab:efficiency_analysis}
    \vspace{-10pt}
\end{table}

As shown in \cref{tab:efficiency_analysis}, TRACE demonstrates a superior balance between computational cost and retrieval performance. On simple tasks like MSCOCO, forcing the model to generate reasoning traces causes a severe performance drop from 87.40\% to 63.90\%, indicating that simple queries suffer from over-thinking and hallucinated constraints. TRACE effectively bypasses this issue, recovering performance to 89.10\% while achieving nearly double the throughput. Conversely, on complex tasks like CIRR, TRACE intelligently trades speed for precision, securing a significant accuracy gain. 

To further validate that TRACE genuinely internalizes task difficulty rather than relying on heuristic guessing or superficial data fitting, we analyzed its routing accuracy on the test set. Remarkably, for simple queries in MSCOCO, the model exhibits a 96\% probability of directly outputting the \texttt{<|emb|>} token at the first decoding step. Conversely, for complex compositional instructions in CIRR, it autonomously chooses to generate text tokens with a 62\% probability. This high routing precision confirms that the adaptive mechanism effectively captures the underlying cognitive demands of diverse queries, achieving a highly favorable and dynamic trade-off.

\subsection{Performance on Universal Retrieval}
\label{sec:mbeir_results}

\textbf{Analysis on the M-BEIR Benchmark.} \Cref{tab:mbeir} demonstrates that TRACE establishes a new state-of-the-art on the comprehensive M-BEIR benchmark. A detailed inspection reveals that performance gains are most pronounced in tasks demanding high-level cognitive processing. On reasoning-intensive datasets such as CIRR~\cite{Liu_2021_ICCV}, FashionIQ~\cite{wu2021fashion}, and InfoSeek~\cite{chen2023infoseek}, TRACE achieves remarkable improvements of 4.2\%, 3.2\%, and 3.8\% in Recall@5 compared to strong baselines like LamRA. 

Traditional universal retrievers typically fail on these datasets because they attempt to map compositional instructions directly into the visual space without intermediate logic. By contrast, TRACE empirically validates that generated reasoning traces effectively bridge this semantic gap. Furthermore, this enhanced capability does not compromise foundational retrieval performance. On standard single-modality tasks, TRACE maintains highly competitive results. The transformative impact of our framework is most evident when comparing TRACE directly to its base model: it propels the vanilla Qwen2.5-VL average score from 23.0\% to 58.8\%. This massive gain confirms that the task-adaptive reasoning paradigm is essential to operationalize the latent knowledge of multimodal models for discriminative tasks.

\input{tables/zeroshot}

\subsection{Scalability Across Backbones}
\label{sec:scalability}

To verify generalizability, we evaluate TRACE across different model sizes and architectures. Empirically, scaling the backbone from Qwen2-VL-2B to Qwen2-VL-7B improves the overall M-BEIR average score from 50.2\% to 55.6\%, and the reasoning-heavy CIRR R@5 from 51.9\% to 54.2\%. Further upgrading to the Qwen2.5-VL-7B architecture yields our best performance, reaching 58.8\% and 57.3\% on the respective metrics. This consistent upward trend confirms that retrieval performance scales directly with the backbone's deductive capacity: stronger models generate higher-quality reasoning traces, which compress into more discriminative embeddings.

\subsection{Zero-Shot Generalization to Unseen Scenarios}
\label{sec:zeroshot_results}

\textbf{Transferability on Unseen Datasets.} 
To strictly validate the transferability of TRACE, we conduct extensive zero-shot experiments on 13 datasets completely isolated from the training phase. \ref{tab:zero-shot} reveals clear advantages. While TRACE performs comparably to massive models like EVA-CLIP-18B~\cite{sun2024eva} on standard matching tasks, it demonstrates exceptional generalization on reasoning heavy benchmarks. 

On CIRCO~\cite{baldrati2023zero}, a highly challenging zero-shot composed retrieval dataset featuring numerous hard distractors, TRACE outperforms LamRA by a clear margin of 1.6\% in MAP@5. Furthermore, on multi-turn interactive tasks such as Multi-round FashionIQ and Visual Dialog~\cite{das2017visual}, we achieve gains of 2.3\% and 2.6\%. We attribute this robust zero-shot performance directly to the internalization of the reasoning process. Instead of memorizing specific dataset distributions, TRACE learns a generalized cognitive skill for intent deconstruction. This enables the model to autonomously adapt to novel retrieval logic and complex semantic constraints in completely unfamiliar domains.

\subsection{Ablation Study}
\label{sec:ablation}

We conduct comprehensive ablation studies on the CIRR subset of our M-BEIR-CoT dataset. It is crucial to note that we utilize the rigorous M-BEIR evaluation protocol~\cite{wei2023uniir}, which retrieves targets from a massive candidate pool of approximately 21K images. This presents a significantly more challenging scenario than the standard CIRR validation split, enforcing strict discrimination against hard negatives. We include more ablation results in the supplementary material.

\textbf{Impact of Feature Extraction Position.}
In causal language models, the hidden state at position $t$ is optimized to predict the token at position $t+1$. Therefore, identifying the optimal position for extracting the retrieval embedding is non-trivial. We compare three extraction strategies: using the hidden state of the \texttt{<|emb|>} token itself, wrapping the reasoning in structural tags and using the final tag, and using the hidden state of the token immediately preceding \texttt{<|emb|>}. \Cref{tab:ablation_pooling} shows that the pre-token strategy achieves the highest performance. From an autoregressive modeling perspective, this hidden state is exactly the one responsible for predicting the sequence-terminating embedding token. Consequently, it acts as a semantic bottleneck, effectively condensing all preceding reasoning steps and visual context into a single representation optimized for retrieval.

\begin{table}[tb]
    \centering
    \caption{\textbf{Ablation on feature extraction position.} The pre-token strategy provides the optimal semantic bottleneck.}
    \begin{tabular}{lccc}
    \toprule
    Pooling Position & R@1 & R@5 & R@10 \\
    \midrule
    Last Token (\texttt{<|emb|>}) & 24.46 & 52.25 & 63.81 \\
    Structure \texttt{</think>} & 24.36 & 53.06 & 63.93 \\
    \textbf{Pre-Token (Ours)} & \textbf{28.37} & \textbf{57.03} & \textbf{68.30} \\
    \bottomrule
    \end{tabular}
    \label{tab:ablation_pooling}
        \vspace{-5pt}
\end{table}

\begin{table}[tb]
    \centering
    \caption{\textbf{Ablation on reasoning data components.} Full reasoning traces yield the highest retrieval accuracy.}
    \begin{tabular}{lccc}
    \toprule
    Data Components & R@1 & R@5 & R@10 \\
    \midrule
    Direct Training (None) & 24.52 & 53.15 & 63.10 \\
    \texttt{<answer>} Only & 25.99 & 53.67 & 64.48 \\
    \texttt{<reasoning>} Only & 26.38 & 53.41 & 64.60 \\
    \textbf{Full CoT} & \textbf{28.37} & \textbf{57.03} & \textbf{68.30} \\
    \bottomrule
    \end{tabular}
    \label{tab:ablation_components}
    \vspace{-15pt}
\end{table}

\textbf{Effectiveness of CoT Components.}
We decompose our reasoning data to analyze which elements contribute most to retrieval accuracy. We train variants using direct mapping without intermediate tokens, using only the final rewritten answer text, using only the logic chain, and using the full reasoning sequence. \Cref{tab:ablation_components} reveals a clear performance hierarchy. Direct training yields the lowest results, confirming that simple black-box mapping is insufficient. While adding isolated answers or logic chains improves over the baseline, utilizing the full trace yields the best results. This demonstrates that the step-by-step process of intent deconstruction combined with an explicit conclusion provides the richest semantic guidance.

\textbf{Comparison with Two-Stage Pipelines.}
To rigorously validate the necessity of end-to-end internalization, we compared TRACE against a decoupled ``External-CoT + Encoder'' baseline. In this two-stage setup, we first utilized the base Qwen2.5-VL model to explicitly generate the reasoning text and target description, and subsequently fed this augmented text into a frozen LamRA encoder. Empirically, the retrieval performance of this disjointed pipeline suffered a severe collapse. This confirms our core hypothesis: compressing the continuous, latent reasoning state directly into the \texttt{<|emb|>} token retains far richer semantic nuances than forcing it through a discrete, hand-off textual bottleneck.

\textbf{Asymmetric Reasoning Strategy.}
We investigate whether generating reasoning is beneficial for both the query and the candidate images. We apply different reasoning configurations to the candidate side during training. \Cref{tab:ablation_asymmetry} reveals a striking phenomenon: applying reasoning to the candidate side causes a catastrophic degradation in performance, plummeting R@5 from 57.03\% to 18.90\%. We attribute this to the fundamental asymmetry of retrieval tasks. The query contains an unresolved user intent that requires active logical deduction to project the representation toward a target state. In contrast, the candidate image acts as a static ground-truth anchor representing the final visual state. Forcing the model to generate text for the candidate causes the embedding to overfit to the generated linguistic patterns rather than capturing the intrinsic visual features, thereby severing the alignment with the query space.

\begin{table}[tb]
    \centering
    \caption{\textbf{Ablation on asymmetric reasoning.} Applying reasoning to the candidate side causes severe performance degradation.}
    \begin{tabular}{llccc}
    \toprule
    Query Side & Candidate Side & R@1 & R@5 & R@10 \\
    \midrule
    Full CoT & Full CoT & 7.75 & 18.90 & 26.50 \\
    Full CoT & \texttt{<answer>} & 8.61 & 19.78 & 26.88 \\
    \textbf{Full CoT} & \textbf{None (Raw)} & \textbf{28.37} & \textbf{57.03} & \textbf{68.30} \\
    \bottomrule
    \end{tabular}
    \label{tab:ablation_asymmetry}
    \vspace{-10pt}
\end{table}

\subsection{Qualitative Visualization}
\label{sec:viz}

\begin{figure}[tb]
  \centering
  \includegraphics[width=\textwidth]{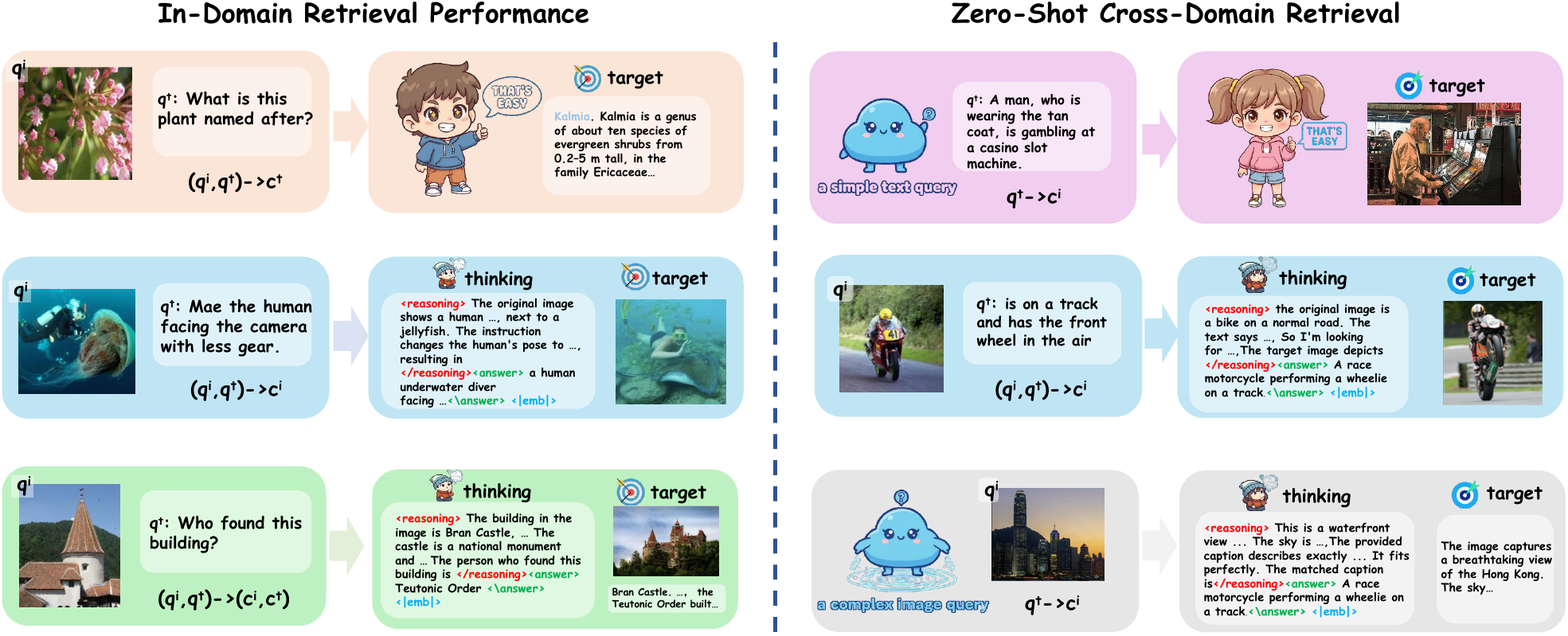}
  \caption{\textbf{Visualization of Adaptive Activation.} \textbf{(Left)} In-Domain Retrieval: TRACE dynamically toggles between a direct path and a reasoning path based on query complexity. \textbf{(Right)} Zero-Shot Generalization: The adaptive behavior effectively transfers to unseen domains and novel constraints.}
  \label{fig:viz_qualitative}
\end{figure}

To provide an intuitive understanding of the framework, we visualize the inference process in Figure~\ref{fig:viz_qualitative}. The visualization explicitly highlights the adaptive activation capability across different data distributions.

In the left panel, TRACE dynamically adjusts its inference depth based on task complexity. For straightforward knowledge recall tasks, the model identifies the low cognitive load and directly generates the embedding token, maximizing efficiency. Conversely, for complex compositional instructions requiring specific object modification, it spontaneously activates the reasoning process to explicitly decompose the visual semantics before compression. 

The right panel demonstrates that this adaptive capability robustly transfers to unseen domains. Even without specific fine-tuning, TRACE autonomously determines when to trigger logical deduction to handle novel constraints or complex state descriptions. This qualitative evidence confirms that the model has acquired a transferable cognitive skill for intent resolution, rather than merely memorizing training set distributions.

\section{Discussion and Broader Impact}
\label{sec:discussion}

\textbf{Limitations.} 
While TRACE significantly advances reasoning-aware retrieval, we acknowledge limitations regarding computational efficiency and data synthesis. First, processing complex intents relies on autoregressive generation, introducing overhead compared to purely feed-forward encoders. However, this trade-off is highly worthwhile, as the ability to handle multi-step reasoning unlocks applications beyond the reach of static models. Second, because M-BEIR-CoT is synthesized via foundation models, the quality of compressed embeddings is intrinsically tied to the teacher model's reasoning ceiling. Consequently, TRACE may inherit specific deductive biases or hallucinate in extreme out-of-distribution scenarios. However, our experiments indicate that these potential biases do not materially affect performance in the regimes we evaluate. Future work will explore speculative decoding to minimize inference latency, alongside human-in-the-loop curation to mitigate dataset biases and enhance reasoning robustness.

\textbf{Broader Societal Impact.} 
The development of universal multimodal retrieval systems capable of complex intent deconstruction has significant positive implications. For instance, it can vastly improve accessibility tools, allowing visually impaired users to search and navigate visual environments using nuanced natural language descriptions. However, we must also consider potential negative societal impacts. If deployed without strict access controls, advanced intent-aware retrieval models could be misused for unauthorized surveillance or the automated extraction of sensitive personal information from massive unstructured databases. We advocate for the responsible development of such technologies, strongly recommending integration with robust privacy-preserving protocols and continuous bias monitoring.

\section{Conclusion} 
In this paper, we introduced TRACE, a novel framework that shifts the paradigm of universal multimodal retrieval from direct encoding to a reasoning then encoding process. By explicitly integrating generative reasoning into the discriminative embedding pipeline, TRACE enables multimodal models to effectively deconstruct complex compositional intents. To support this capability, we constructed M-BEIR-CoT, a large-scale dataset synthesized via a difficulty-aware pipeline. Crucially, we introduced a learned adaptive mechanism that resolves the inherent tension between reasoning depth and inference efficiency. This allows the model to spontaneously bypass explicit reasoning for simple queries while activating it for complex instructions. Extensive experiments demonstrate that TRACE establishes new state-of-the-art performance across the M-BEIR benchmark and various zero-shot scenarios. Furthermore, our analysis uncovers a fundamental asymmetry in retrieval logic, proving that reasoning is exclusively beneficial on the query side. We anticipate that our framework will provide a robust foundation for building more interpretable and cognitively advanced retrieval systems. 
\clearpage

%
%
\bibliographystyle{splncs04}
\bibliography{main}

\input{sec/X_suppl}
\end{document}

%% file: tables/mbeir.tex
\begin{table}[tb]
\centering
\caption{\textbf{Comparison on the M-BEIR test set.} We report Recall@5 for the majority of datasets, with the exception of FashionIQ (FIQ)~\cite{wu2021fashion} and Fashion200K (F200K)~\cite{han2017automatic} where Recall@10 is utilized. The first row indicates the specific retrieval task type (e.g. $q^t \to c^i$ for text-to-image retrieval). Abbreviations: VN (VisualNews)~\cite{liu2020visualnews}, InfoS (InfoSeek)~\cite{chen2023infoseek}. The best results are highlighted in \textbf{bold}, and the second-best are \underline{underlined}. The \textit{Improvement} row specifically reports the performance gains of TRACE compared to LamRA-Ret.}
\definecolor{dspgreen}{rgb}{0.0, 0.6, 0.0}
\resizebox{\linewidth}{!}{
\begin{tabular}{lc@{\hspace{0.1cm}}c@{\hspace{0.1cm}}c@{\hspace{0.1cm}}c@{\hspace{0.1cm}}c@{\hspace{0.1cm}}c@{\hspace{0.1cm}}c@{\hspace{0.1cm}}c@{\hspace{0.1cm}}c@{\hspace{0.1cm}}c@{\hspace{0.1cm}}c@{\hspace{0.1cm}}c@{\hspace{0.1cm}}c@{\hspace{0.1cm}}c@{\hspace{0.1cm}}c@{\hspace{0.1cm}}c@{\hspace{0.1cm}}c@{\hspace{0.1cm}}}
\toprule
 & \multicolumn{3}{c}{{$q^t \to c^i$}} & {$q^t \to c^t$} & \multicolumn{2}{c}{{$q^t \to (c^i, c^t)$}} & \multicolumn{3}{c}{{$q^i \to c^t$}} & {$q^i \to c^i$} & \multicolumn{2}{c}{{$(q^i, q^t) \to c^t$}} & \multicolumn{2}{c}{{$(q^i, q^t) \to c^i$}} & \multicolumn{2}{c}{{$(q^i, q^t) \to (c^i, c^t)$}} & \\
 \cmidrule(r){2-4} \cmidrule(r){5-5}  \cmidrule(r){6-7} \cmidrule(r){8-10} \cmidrule(r){11-11} \cmidrule(r){12-13} \cmidrule(r){14-15} \cmidrule(r){16-17} 
 Methods & VN  & COCO & F200K & WebQA & EDIS & WebQA & VN & COCO & F200K & NIGHTS & OVEN & InfoS & FIQ & CIRR & OVEN & InfoS & Avg. \\
\cmidrule(r){2-4} \cmidrule(r){5-5}  \cmidrule(r){6-7} \cmidrule(r){8-10} \cmidrule(r){11-11} \cmidrule(r){12-13} \cmidrule(r){14-15} \cmidrule(r){16-17} 
& R@5 & R@5 & R@10 & R@5 & R@5 & R@5 & R@5 & R@5 & R@10 & R@5 & R@5 & R@5 & R@10 & R@5 & R@5 & R@5 & \\
\midrule
CLIP-L~\cite{radford2021learning} & \textbf{43.3} & 61.1 & 6.6 & 36.2 & 43.3 & 45.1 & \underline{41.3} & 79.0 & 7.7 & 26.1 & 24.2 & 20.5 & 7.0 & 13.2 & 38.8 & 26.4 & 32.5 \\
SigLIP~\cite{zhai2023sigmoid} & 30.1 & 75.7 & \textbf{36.5} & 39.8 & 27.0 & 43.5 & 30.8 & 88.2 & \textbf{34.2} & 28.9 & 29.7 & 25.1 & 14.4 & 22.7 & 41.7 & 27.4 & 37.2 \\
BLIP~\cite{li2022blip} & 16.4 & 74.4 & 15.9 & 44.9 & 26.8 & 20.3 & 17.2 & 83.2 & 19.9 & 27.4 & 16.1 & 10.2 & 2.3 & 10.6 & 27.4 & 16.6 & 26.8 \\
BLIP2~\cite{li2023blip} & 16.7 & 63.8 & 14.0 & 38.6 & 26.9 & 24.5 & 15.0 & 80.0 & 14.2 & 25.4 & 12.2 & 5.5 & 4.4 & 11.8 & 27.3 & 15.8 & 24.8 \\
Qwen2.5-VL-7B~\cite{qwen2.5vl} & 9.3 & 55.1 & 5.0 & 42.0 & 26.2 & 9.4 & 5.4 & 46.6 & 4.0 & 21.3 & 21.4 & 22.5 & 4.3 & 16.3 & 43.6 & 36.2 & 23.0 \\
$\text{UniIR-BLIP}_{\text{FF}}$~\cite{wei2023uniir} & 23.4 & 79.7 & 26.1 & 80.0 & 50.9 & 79.8 & 22.8 & 89.9 & 28.9 & \underline{33.0} & 41.0 & 22.4 & 29.2 & 52.2 & 55.8 & 33.0 & 46.8 \\
$\text{UniIR-CLIP}_{\text{SF}}$~\cite{wei2023uniir} & \underline{42.6} & 81.1 & 18.0 & 84.7 & 59.4 & 78.7 & \textbf{43.1} & \textbf{92.3} & 18.3 & 32.0 & 45.5 & 27.9 & 24.4 & 44.6 & 67.6 & 48.9 & 50.6 \\
LamRA-Ret~\cite{liu2025lamra} & 41.6 & \underline{81.5} & 28.7 & \underline{86.0} & \underline{62.6} & \underline{81.2} & 39.6 & 90.6 & 30.4 & 32.1 & \underline{54.1} & \underline{52.1} & \underline{33.2} & \underline{53.1} & \underline{76.2} & \underline{63.3} & \underline{56.6} \\
\midrule
\multicolumn{18}{c}{\textit{Our Method}} \\
\midrule
TRACE & 42.1 & \textbf{82.3} & \underline{30.5} & \textbf{87.8} & \textbf{64.1} & \textbf{82.5} & 41.2 & \underline{91.3} & \underline{33.2} & \textbf{33.6} & \textbf{57.4} & \textbf{55.8} & \textbf{36.4} & \textbf{57.3} & \textbf{78.5} & \textbf{67.1} & \textbf{58.8} \\
\textit{- Improvement} & \textcolor{dspgreen}{\textit{+0.5}} & \textcolor{dspgreen}{\textit{+0.8}} & \textcolor{dspgreen}{\textit{+1.8}} & \textcolor{dspgreen}{\textit{+1.8}} & \textcolor{dspgreen}{\textit{+1.5}} & \textcolor{dspgreen}{\textit{+1.3}} & \textcolor{dspgreen}{\textit{+1.6}} & \textcolor{dspgreen}{\textit{+0.7}} & \textcolor{dspgreen}{\textit{+2.8}} & \textcolor{dspgreen}{\textit{+1.5}} & \textcolor{dspgreen}{\textit{+3.3}} & \textcolor{dspgreen}{\textit{+3.7}} & \textcolor{dspgreen}{\textit{+3.2}} & \textcolor{dspgreen}{\textit{+4.2}} & \textcolor{dspgreen}{\textit{+2.3}} & \textcolor{dspgreen}{\textit{+3.8}} & \textcolor{dspgreen}{\textit{+2.2}} \\
\bottomrule
 \end{tabular}
}

\label{tab:mbeir}
\end{table}

%% file: tables/zeroshot.tex
\begin{table}[tb]
\setlength{\tabcolsep}{1.5mm}
\definecolor{dspgreen}{rgb}{0.0, 0.6, 0.0}
\centering
\caption{\textbf{Comparison with unseen datasets (Zero-shot).} The columns indicate task types, covering text query ($q^t$), image query ($q^i$), dialog query ($q^{\text{dialog}}$), and multi-interleaved queries.  Abbreviations: Share4V (ShareGPT4V)~\cite{chen2023sharegpt4v}, Urban (Urban-1k)~\cite{zhang2024long}, VIST (Visual Storytelling)~\cite{dong2025vistavisualstorytellingusing}, VisD (Visual Dialog)~\cite{das2017visual}, and MT-FIQ (Multi-round FashionIQ)\cite{wu2021fashion}. $*$ indicates domain shifts from training data. We report R@1 for most datasets, MAP@5 for CIRCO~\cite{baldrati2023zero}, R@5 for MT-FIQ, and Accuracy for ITM tasks. The best results are highlighted in \textbf{bold}, and the second-best are \underline{underlined}. The \textit{Improvement} row reports the performance gain of TRACE specifically compared to LamRA-Ret.} 
\resizebox{\linewidth}{!}{
\begin{tabular}{lccccccccccccc}
\toprule
 & \multicolumn{3}{c}{{$q^t \to c^i$}} & \multicolumn{3}{c}{{$q^i \to c^t$}} & \multicolumn{2}{c}{{$(q^i, q^t) \to c^i$}} & \multicolumn{1}{c}{{$q^{\text{dialog}} \to c^i$}} & 
\multicolumn{2}{c}{{$(q^i \oplus q^t) \to c^i$}} &\multicolumn{2}{c}{{ITM}}\\
\cmidrule(r){2-4} \cmidrule(r){5-7}  \cmidrule(r){8-9} \cmidrule(r){10-10}  \cmidrule(r){11-12} \cmidrule(r){13-14} 
Methods & Share4V  & Urban$^*$ & Flickr & Share4V  & Urban$^*$ & Flickr  & CIRCO$^*$ & GeneCIS$^*$ & VisD$^*$ & VIST & MT-FIQ$^*$ & CC-Neg & Sugar-Crepe$^*$ \\
\cmidrule(r){2-4} \cmidrule(r){5-7}  \cmidrule(r){8-9} \cmidrule(r){10-10}  \cmidrule(r){11-12} \cmidrule(r){13-14} 
& R@1 & R@1 & R@1 & R@1 & R@1 & R@1 & MAP@5 & R@1 & R@1 & R@1 & R@5 & Acc. & Acc. \\
\midrule
CLIP-L~\cite{radford2021learning} & 84.0 & 52.8 & 67.3 & 81.8 & 68.7 & 87.2 & 4.0 & 13.3 & 23.7 & 0.6 & 17.7 & 66.7 & 73.0 \\
Long-CLIP-L~\cite{zhang2024long} & \underline{95.6} & 86.1 & 76.1 & \underline{95.8} & 82.7 & 89.3 & 5.7 & 16.3 & 37.9 & 1.1 & 18.5 & 76.3 & 80.9 \\
UniIR-CLIP~\cite{wei2023uniir} & 85.8 & 75.0 & 78.7 & 84.1 & 78.4 & 94.2 & 12.5 & 16.8 & 26.8 & 0.6 & 39.4 & 79.9 & 80.3 \\
E5-V~\cite{jiang2024e5} & 86.7 & 84.0 & 79.5 & 84.0 & 82.4 & 88.2 & 24.8 & 18.5 & 54.6 & 10.0 & 19.2 & \underline{83.2} & 84.7 \\
MagicLens-L~\cite{zhang2024magiclens} & 85.5 & 59.3 & 72.5 & 60.9 & 24.2 & 84.6 & 29.6 & 16.3 & 28.0 & 3.3 & 22.6 & 62.7 & 75.9 \\
EVA-CLIP-8B~\cite{sun2024eva} & 91.2 & 77.8 & 80.8 & 93.1 & 80.4 & 95.6 & 6.0 & 13.1 & 23.2 & 1.2 & 22.1 & 59.4 & 81.7\\
EVA-CLIP-18B~\cite{sun2024eva} & 92.1 & 81.7 & \underline{83.3} & 94.0 & 83.3 & \textbf{96.7} & 6.1 & 13.6 & 24.7 & 1.0 & 21.9 & 63.8 & 83.1\\
LamRA-Ret~\cite{liu2025lamra} & 93.3 & \textbf{95.1} & 82.8 & 88.1  & \textbf{94.3} & 92.7 & \textbf{33.2} & \textbf{18.9}  & \textbf{62.8} & \textbf{23.1} & \textbf{60.9} & 79.6 & \textbf{85.8} \\
\midrule
\multicolumn{14}{c}{\textit{Ours Method}} \\
\midrule
TRACE & \textbf{94.9} & \underline{94.8} & \textbf{84.5} & \textbf{89.1} & \underline{94.1} & \underline{94.5} & \underline{34.8} & \underline{20.5} & \underline{65.4} & \underline{25.8} & \underline{63.2} & \textbf{84.1} & \underline{87.5} \\
\textit{- Improvement} & \textcolor{dspgreen}{\textit{+1.6}} & \textcolor{dspgreen}{\textit{-0.3}} & \textcolor{dspgreen}{\textit{+1.7}} & \textcolor{dspgreen}{\textit{+1.0}} & \textcolor{dspgreen}{\textit{-0.2}} & \textcolor{dspgreen}{\textit{+1.8}} & \textcolor{dspgreen}{\textit{+1.6}} & \textcolor{dspgreen}{\textit{+1.6}} & \textcolor{dspgreen}{\textit{+2.6}} & \textcolor{dspgreen}{\textit{+2.7}} & \textcolor{dspgreen}{\textit{+2.3}} & \textcolor{dspgreen}{\textit{+4.5}} & \textcolor{dspgreen}{\textit{+1.7}} \\
\bottomrule
 \end{tabular}
 
}

\label{tab:zero-shot}
\end{table}

%% file: sec/X_suppl.tex
\clearpage
\setcounter{page}{1}

\makeatletter
\begin{center}
    \Large
    \textbf{TRACE: Task-Adaptive Reasoning and Representation Learning for Universal Multimodal Retrieval}\\
    \vspace{0.5em}
    Supplementary Material\\
    \vspace{1.0em}
\end{center}
\makeatother

\section{More Implementation Details}
In this section, we provide further details regarding our experimental setup, including the specific prompts used for task-adaptive reasoning, ablations on LoRA configurations, and the analysis of loss weighting.

\begin{figure*}[t]
    \centering
    \begin{subfigure}[b]{0.9\textwidth}
        \centering
        \includegraphics[width=\textwidth]{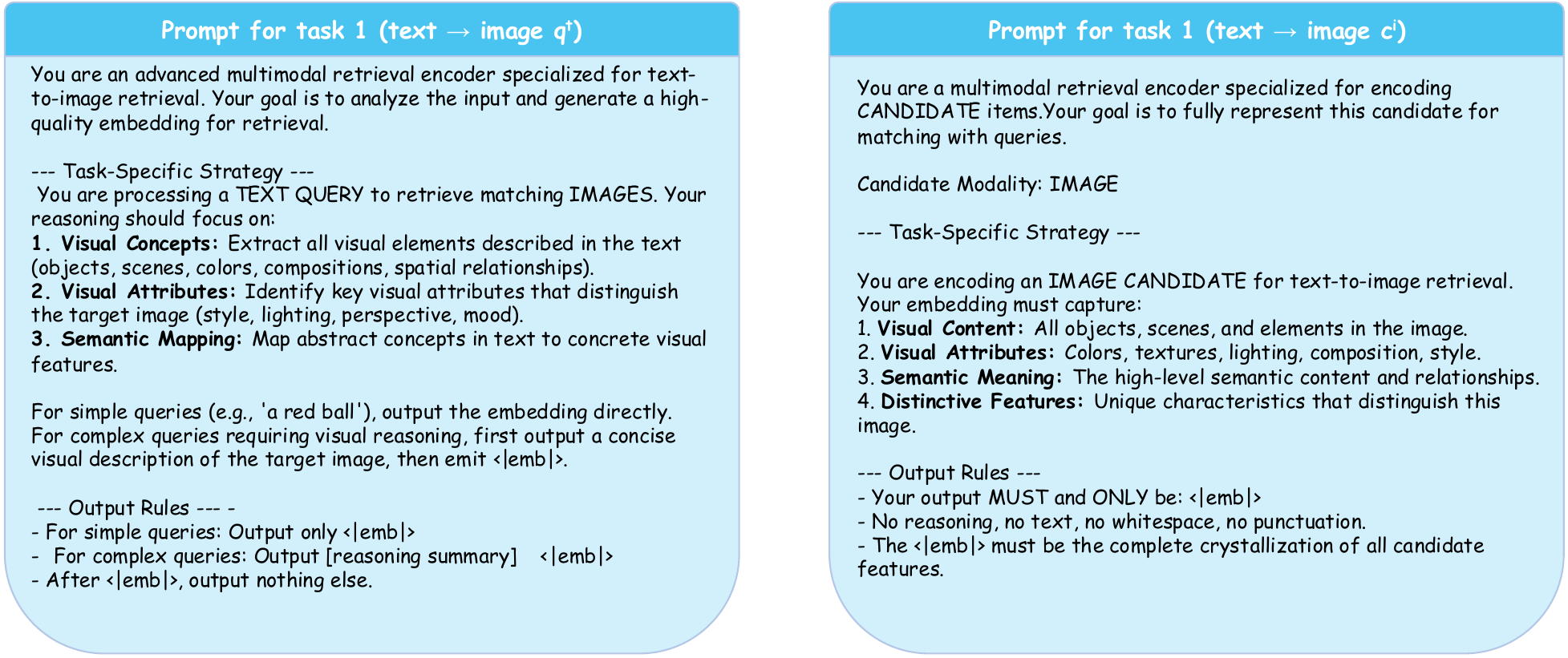}
        \caption{\textbf{Prompt for Text-to-Image Retrieval.}}
        \label{fig:prompt_t2i}
    \end{subfigure}
    
    \vspace{1.5em}
    \begin{subfigure}[b]{0.9\textwidth}
        \centering
        \includegraphics[width=\textwidth]{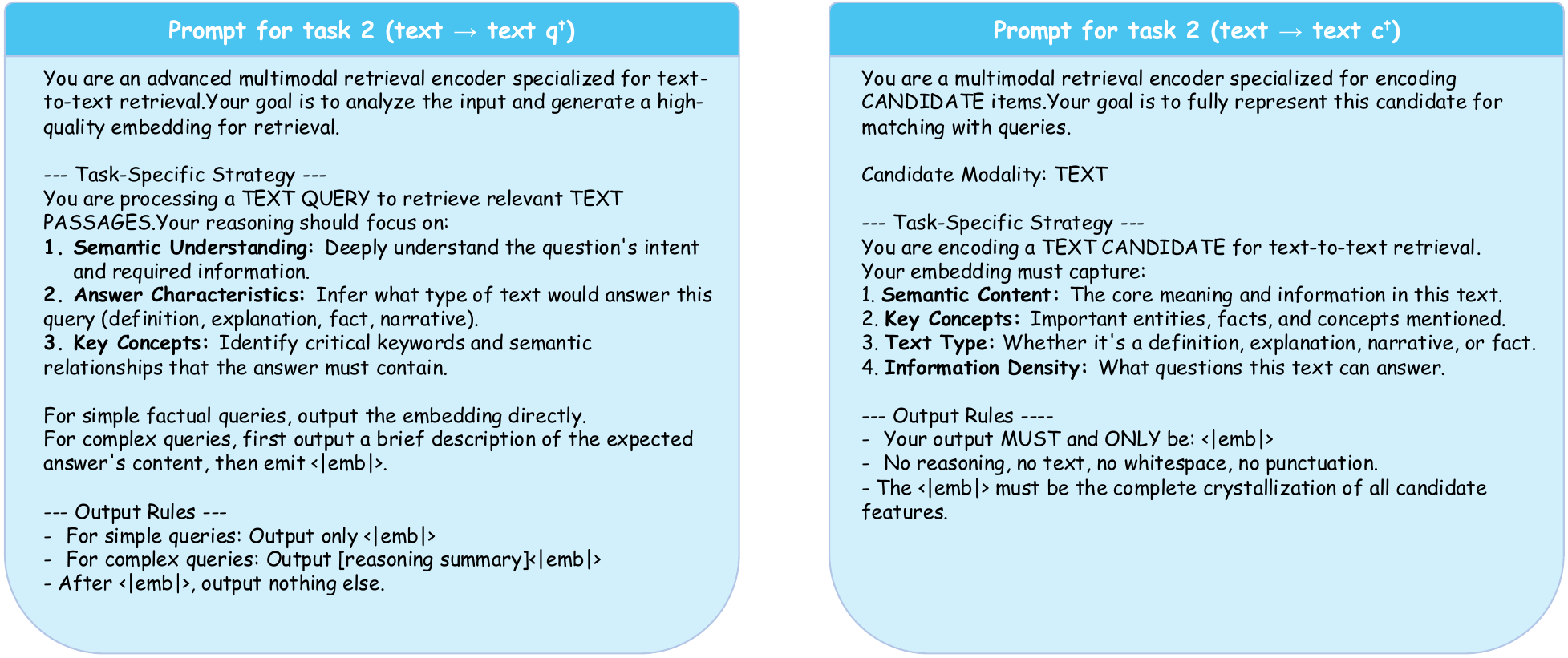}
        \caption{\textbf{Prompt for Text-to-Text Retrieval.}}
        \label{fig:prompt_cirr}
    \end{subfigure}

    \vspace{1.5em}
    \begin{subfigure}[b]{0.9\textwidth}
        \centering
        \includegraphics[width=\textwidth]{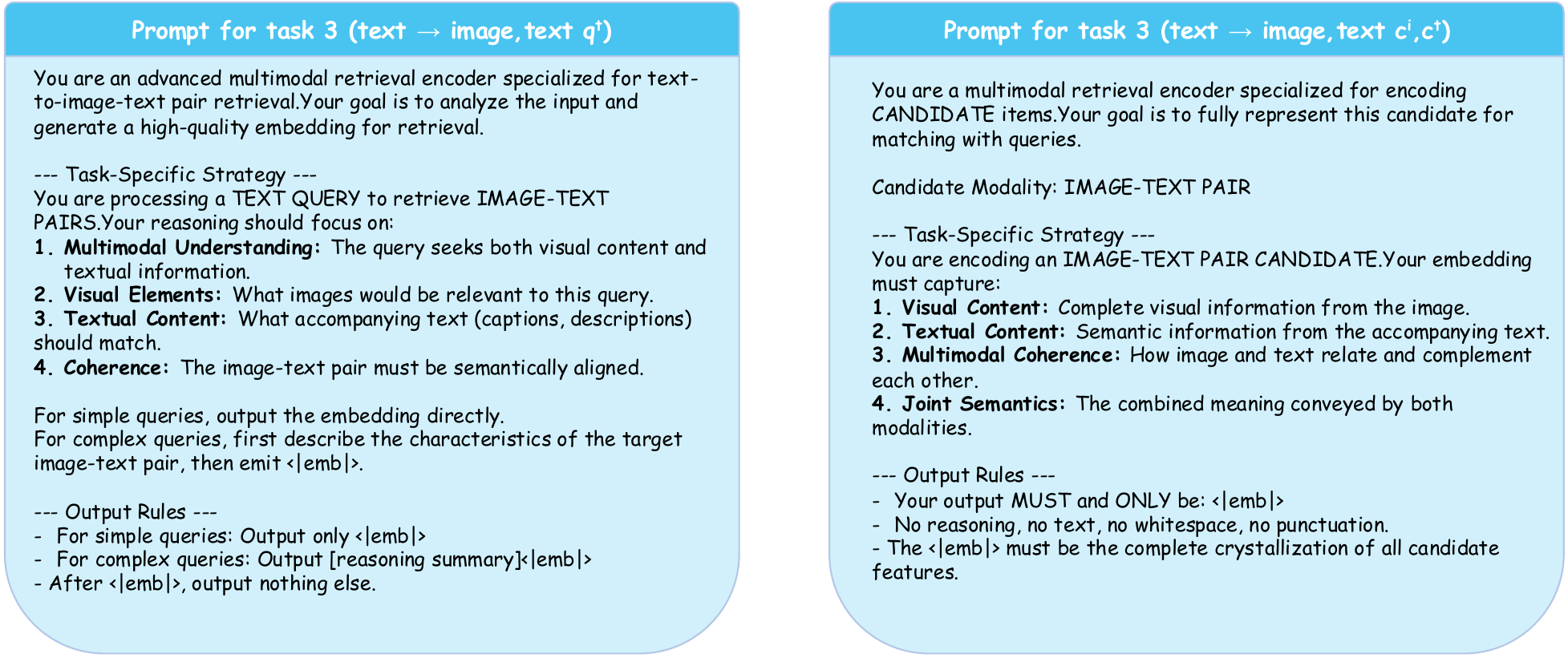}
        \caption{\textbf{Prompt for Text-to-(Image,Text) Retrieval.}}
        \label{fig:prompt_vqa}
    \end{subfigure}

    \caption{\textbf{Examples of task-specific prompts used in M-BEIR-CoT.}}
    \label{fig:prompt_examples}
\end{figure*}

\noindent \textbf{Task-Specific Prompts.}
Our \textbf{M-BEIR-CoT} dataset is designed to teach the model to differentiate task intent. To achieve this, we prefix the input with a task-specific instruction prompt. These prompts are crucial for guiding the MLLM to understand the specific intent of each retrieval task (e.g., Composed Retrieval vs. VQA) and to trigger the correct reasoning path. We illustrate some of the key prompt templates used in Figure~\ref{fig:prompt_examples}.

\noindent \textbf{Ablation on Fine-tuning Strategy.}
As discussed in the main paper (\cref{sec:ablation}), we use Low-Rank Adaptation (LoRA) for parameter-efficient fine-tuning, which enables larger batch sizes. To determine the optimal LoRA hyperparameters, we conducted experiments on the \textbf{CIRR} subset (as defined in \cref{sec:ablation}) by varying the rank (\textit{r}) and scale (\textit{alpha}).

As shown in Table~\ref{tab:lora_ablation}, we found that increasing the rank generally improves performance, but with diminishing returns. The configuration of \textbf{r=128} and \textbf{alpha=256} (a common 2:1 ratio) achieved the highest performance. While \textit{r=256} was comparable, it introduced a higher parameter overhead, making \textit{r=128} the optimal choice for our final model.

\input{supp_tables/lora_ablation}

\noindent \textbf{Ablation on Loss Weights.}
Our total objective is a weighted sum of the generative loss and the contrastive retrieval loss:
$\mathcal{L} = \lambda_{\text{gen}} \mathcal{L}_{\text{gen}} + \lambda_{\text{ret}} \mathcal{L}_{\text{ret}}$.
To find the optimal balance, we fixed $\lambda_{\text{ret}}=1.0$ and varied the weight of the generative reasoning loss ($\lambda_{\text{gen}}$) on the CIRR subset.

The results are presented in Table~\ref{tab:loss_ablation}.
We observe that without the generative loss ($\lambda_{\text{gen}}=0$), the model fails to learn the reasoning process, leading to poor retrieval performance. Conversely, assigning too much weight to generation ($\lambda_{\text{gen}}=2.0$) distracts the model from the primary retrieval objective, also degrading performance. A 1:1 weighting ($\lambda_{\text{gen}}=1.0$) provided the best balance, ensuring that the reasoning process is explicitly guided to support the discriminative retrieval task.

\input{supp_tables/loss_ablation}

\section{Further Analysis}

As shown in the main paper (\cref{tab:ablation_asymmetry}), applying CoT to the candidate side leads to a catastrophic performance degradation. A natural question arises: why does this happen, given that the \textit{candidate-side prompts} also have variable lengths, yet do not cause such a collapse?

We hypothesize that this discrepancy is mainly due to a qualitative difference in the \textbf{nature and magnitude} of the positional shifts introduced by short prompts versus those introduced by generated CoT, together with the semantic role that the candidate embedding is supposed to play in our contrastive space.

\textbf{Understanding RoPE and Relative Distance.}
Our backbone (Qwen2.5-VL) uses Rotary Position Embeddings (RoPE), which do not assign an absolute position to each token, but instead encode the \textbf{relative distance} $d$ between any two tokens (e.g., the query token $q_m$ and key token $k_n$) within the attention mechanism. The final representation is extracted from the \texttt{[pre-token]}, so we are primarily concerned with the relative distance $d$ between this \texttt{[pre-token]} and the \texttt{[Image\_c]} tokens associated with a candidate image $I_c$.

\textbf{Benign Case: Candidate-Side Prompts.}
In the standard setup, the candidate input takes the form
\[
\texttt{[Image\_c] [Prompt\_c] [pre-token] [emb]}.
\]
Here, the length of \texttt{[Prompt\_c]} ($L_p$) varies only across a small number of predefined task templates (e.g., ``Describe this image'' vs. ``Answer the question''). This variation is therefore relatively small and structured. In practice, the relative distance $d$ between \texttt{[pre-token]} and \texttt{[Image\_c]} fluctuates only within a narrow range (e.g., an illustrative range of $5 \le d \le 15$ tokens). RoPE is known to generalize well to relative distances that are sufficiently covered during training, and the model repeatedly encounters these few positional configurations when mapping candidates into the embedding space. As a result, it can learn to produce a consistent representation for $I_c$ regardless of which short prompt is used on the candidate side, and candidate-side prompts do not destabilize the contrastive space.

\textbf{Catastrophic Case: Candidate-Side CoT.}
When we force CoT generation on the candidate side, the input becomes
\[
\texttt{[Image\_c] [CoT\_c] [pre-token] [emb]}.
\]
In this case, the length of the generated \texttt{[CoT\_c]} ($L_c$) is stochastic and, in practice, much less controlled than the hand-designed prompts. It can range from only a few tokens to well over a hundred tokens, even for the same image $I_c$. This makes the relative distance
\[
d = L_{\text{Img}} + L_c + 1
\]
highly variable and potentially very large. Under this regime, the model is now asked to aggregate information from \texttt{[Image\_c]} at relative distances that are rarely or never seen when learning the candidate embedding. Intuitively, the attention patterns induced by RoPE at $d \approx 120$ can differ significantly from those around the typical training range (e.g., $d \approx 10$), leading to an out-of-distribution positional configuration for extracting the candidate representation. This makes the resulting embedding $\mathbf{e}'_c$ much less stable across different CoT realizations.

Beyond positional effects, \texttt{[CoT\_c]} itself carries high-entropy semantic content. The candidate embedding is no longer a predominantly visual ``anchor'' for $I_c$, but becomes heavily entangled with the self-generated explanation. This semantic role shift moves $\mathbf{e}'_c$ away from the intended role of a stable, image-grounded target in the contrastive space, and instead makes it sensitive to the particular wording and length of the generated CoT.

Taken together, these observations provide a plausible explanation for the drastic degradation observed when applying CoT on the candidate side. The candidate embedding is designed to serve as a stable, visually grounded anchor $\mathbf{e}_c$ in the contrastive space. While small, deterministic positional shifts from short prompts are easily absorbed by RoPE and the training distribution, the combination of (i) stochastic, potentially large positional shifts induced by long CoT and (ii) semantic contamination from candidate-side reasoning yields a highly out-of-distribution and unstable target. We conjecture that this compounded effect makes it extremely difficult for the model to maintain representation consistency, ultimately leading to the collapse reported in Table~\ref{tab:ablation_asymmetry}. A more systematic study of RoPE behavior under such asymmetric reasoning patterns is left for future work.

\input{supp_tables/simple_subset}
\input{supp_tables/reasoning_subset}

\input{supp_tables/unseen}

\section{Details about M-BEIR-CoT Dataset}
\label{sec:M-BEIR-CoT}

We detail the composition of the M-BEIR-CoT dataset in Table~\ref{tab:simple_subset} and Table~\ref{tab:reasoning_subset}, respectively. This dataset comprises two distinct subsets: the Direct Retrieval Subset (Simple Subset) and the Reasoning-Enhanced Subset (Reasoning Subset).

Simple Subset (Table~\ref{tab:simple_subset}): To maintain fundamental multimodal retrieval capabilities, we directly adopted the majority of data from the M-BEIR benchmark~\cite{wei2023uniir}. Specifically, we retained datasets that rely on pattern matching rather than complex reasoning. As shown in Table~\ref{tab:simple_subset}, this component includes datasets such as VisualNews~\cite{liu2020visualnews}, MSCOCO~\cite{lin2014microsoft}, Fashion200K~\cite{han2017automatic}, WebQA~\cite{chang2022webqamultihopmultimodalqa}, EDIS~\cite{liu2023edisentitydrivenimagesearch}, and NIGHTS~\cite{fu2023dreamsimlearningnewdimensions}. Consistent with the original M-BEIR paper, the training and evaluation splits for these data remain unchanged to ensure experimental comparability.

Reasoning Subset (Table~\ref{tab:reasoning_subset}): For tasks requiring multi-step logical deduction or fine-grained attribute understanding, we constructed specialized Chain-of-Thought (CoT) data. This primarily covers four datasets: CIRR~\cite{Liu_2021_ICCV}, FashionIQ~\cite{wu2021fashion}, InfoSeek~\cite{chen2023infoseek}, and OVEN~\cite{hu2023opendomainvisualentityrecognition}. We employed a rigorous ``generation-filtering'' pipeline to ensure data quality. As indicated by the statistics in Table~\ref{tab:reasoning_subset}, we first generated a large volume of Original CoT data, which was subsequently cleaned via a dual filtering mechanism based on rules and models. Ultimately, 24,935 high-quality samples were retained for CIRR, 15,626 for FashionIQ, while InfoSeek and OVEN retained 241,006 and 293,875 samples, respectively.

\noindent \textbf{Dual Filtering \& Curation Details.}
As illustrated in Fig.~\ref{fig:data_pipeline}, we adopt a coarse-to-fine filtration protocol to remove hallucinated or weakly grounded CoT traces. Given a query-target pair $(Q, C^{+})$ from the \emph{training split} of M-BEIR, we first prompt a strong MLLM to generate a structured response in the form \texttt{<reasoning> ... </reasoning> <answer> ... </answer>}. We use the auxiliary tags only for reliable parsing; during TRACE training, we remove \texttt{<reasoning>} and \texttt{<answer>} tags while keeping the natural language content (as described in \cref{sec:data_construction}).

\noindent \emph{Rule-based filtering (fast, high-recall).} We discard a sample if it violates any of the following rules (corresponding to the ``format/length/keyword'' checks in the right panel of Fig.~\ref{fig:data_pipeline}):
\begin{itemize}
    \item \textbf{Format check.} The output must contain exactly one pair of \texttt{<reasoning>} and \texttt{<answer>} tags, with \texttt{<reasoning>} appearing before \texttt{<answer>}. We reject malformed tags, nested tags, missing closing tags, or obvious prompt leakage/meta tokens (e.g., ``\texttt{assistant:}'', ``\texttt{system:}'').
    \item \textbf{Length limit.} We remove near-empty traces and overly verbose generations to control both supervision quality and training cost. In practice, we require the reasoning length to be within $[16, 256]$ tokens and the answer length within $[2, 64]$ tokens (measured by the teacher tokenizer), and we drop samples with excessive repetition (e.g., repeated $n$-grams) or non-English outputs.
    \item \textbf{Keyword/sanity check.} We filter out refusals and boilerplate (e.g., ``As an AI...'', ``I can't determine...''), and require the \texttt{<answer>} to contain at least one salient keyword from the query instruction or the ground-truth target text (after stopword removal). For composed image retrieval, we additionally remove degenerate cases where the edited answer is identical (after normalization) to the source caption, indicating a failure to apply the modification.
\end{itemize}
We also perform lightweight deduplication by normalizing \texttt{<answer>} (lowercasing and stripping punctuation) and removing near-duplicate (query, answer) pairs.

\noindent \emph{Model-based filtering (slow, high-precision).} For the remaining candidates, we apply a semantic consistency check using a strong verifier model. The verifier outputs a binary decision along with a confidence score in $[0,1]$, and we retain samples only when the decision is \emph{consistent} and the confidence exceeds a conservative threshold ($\tau=0.7$).
For image-target tasks ($ (q^i,q^t)\!\rightarrow\! c^i$, i.e., CIRR/FashionIQ), the verifier is given $(q^i,q^t)$, the ground-truth target image $c^i$, and the generated \texttt{<answer>}, and is asked to judge whether the answer faithfully describes $c^i$ \emph{while satisfying the modification instruction} (e.g., applying the requested change and preserving the invariant context).
For text-target tasks ($ (q^i,q^t)\!\rightarrow\! c^t$, i.e., InfoSeek/OVEN), we verify that the answer refers to the correct target document/entity by checking entailment against the ground-truth title/summary and rejecting vague descriptions that could match multiple candidates.
This dual filtering \& curation step is critical for ensuring that the final 575,442 reasoning samples provide supportive (rather than hallucinatory) supervision.

Furthermore, to ensure fairness when evaluating on the M-BEIR benchmark, we utilized the same instructions (Table~\ref{tab:instruction}) as LamRA~\cite{liu2025lamra} during testing.

\input{supp_tables/instructions}

\section{Details about Unseen Dataset}

Here, we present the details of the Unseen Dataset in Table~\ref{tab:unseen_dataset}. While several datasets originate from MSCOCO or FashionIQ, they feature distinct captioning styles or query formats that fundamentally alter the retrieval dynamics. For instance, Urban1K utilizes extended captions generated by GPT-4V~\cite{openai2023gpt4v}, whereas CIRCO employs a complex query format combining a reference image with a relative caption. Given these substantial distributional shifts compared to the original training data, we categorize them as unseen datasets to strictly evaluate generalization capability.

\section{More Qualitative Results}
\label{sec:supp_qualitative}

\begin{figure*}[t]
    \centering
    \includegraphics[width=1.0\textwidth]{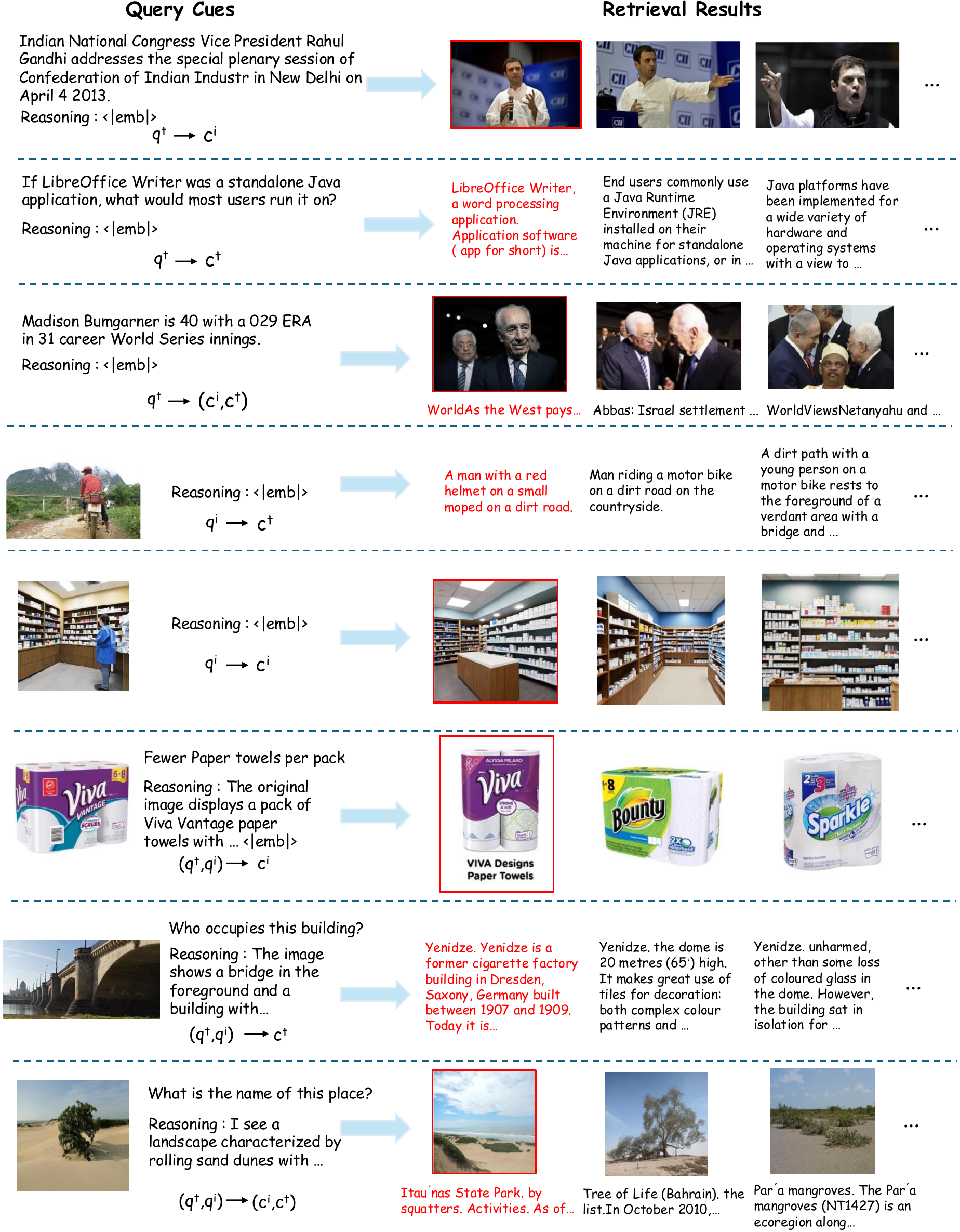}
    \caption{
        \textbf{Qualitative visualization of Universal Multimodal Retrieval tasks performed by TRACE.}
    }
    \label{fig:universal_vis}
\end{figure*}

In this section, we present comprehensive qualitative examples to demonstrate the versatility and universality of our TRACE framework. As illustrated in Figure~\ref{fig:universal_vis}, our model effectively unifies diverse retrieval tasks within a single generative interface, handling arbitrary combinations of query and candidate modalities.

%% file: supp_tables/lora_ablation.tex
\begin{table}[h]
    \centering
    \small
        \caption{\textbf{Ablation on LoRA hyperparameters (r and alpha).}}
    \setlength{\tabcolsep}{10pt} 
    \begin{tabular}{cc|ccc}
    \toprule
    \textbf{Rank (r)} & \textbf{Alpha ($\alpha$)} & \textbf{R@1} & \textbf{R@5} & \textbf{R@10} \\
    \midrule
    16 & 32 & 26.67 & 56.31 & 66.45 \\
    32 & 64 & 26.9 & 56.19 & 67.29 \\
    \textbf{128} & \textbf{256} & \textbf{28.37} & \textbf{57.03} & \textbf{68.30} \\
    128 & 512 & 27.48 & 56.74 & 67.84 \\
    256 & 512 & 28.19 & 56.81 & 68.02 \\
    \bottomrule
    \end{tabular}
    \vspace{-5pt}
    \label{tab:lora_ablation}
\end{table}

%% file: supp_tables/loss_ablation.tex
\begin{table}[h]
    \centering
    \small
    \caption{\textbf{Ablation on loss weights.}}
    \setlength{\tabcolsep}{10pt}
    \begin{tabular}{c|ccc}
    \toprule
    $\lambda_{\text{gen}}$ & R@1 & R@5 & R@10 \\
    \midrule
    0.0 & 24.52 & 53.15 & 63.10 \\
    0.5 & 26.89 & 55.92 & 66.80 \\
    \textbf{1.0} & \textbf{28.37} & \textbf{57.03} & \textbf{68.30} \\
    2.0 & 27.01 & 55.74 & 66.91 \\
    \bottomrule
    \end{tabular}
    \vspace{-5pt}
    \label{tab:loss_ablation}
\end{table}

%% file: supp_tables/simple_subset.tex
\begin{table*}[t]
    \centering
    \small
    \caption{Summary of the \textbf{Simple Subset} in M-BEIR-CoT.}
    \label{tab:simple_subset}
    \begin{tabular}{lllrrrr}
        \toprule
        Task & Dataset & Domain & \# Train & \# Dev & \# Test & \# Pool \\
        \midrule
        \multirow{3}{*}{$q^t \rightarrow c^i$} 
         & VisualNews & News & 99K & 20K & 20K & 542K \\
         & MSCOCO & Misc. & 100K & 24.8K & 24.8K & 5K \\
         & Fashion200K & Fashion & 15K & 1.7K & 1.7K & 201K \\
        \midrule
        $q^t \rightarrow c^t$ & WebQA & Wiki & 16K & 1.7K & 2.4K & 544K \\
        \midrule
        \multirow{2}{*}{$q^t \rightarrow (c^i, c^t)$} 
         & EDIS & News & 26K & 3.2K & 3.2K & 1M \\
         & WebQA & Wiki & 17K & 1.7K & 2.5K & 403K \\
        \midrule
        \multirow{3}{*}{$q^i \rightarrow c^t$} 
         & VisualNews & News & 100K & 20K & 20K & 537K \\
         & MSCOCO & Misc. & 113K & 5K & 5K & 25K \\
         & Fashion200K & Fashion & 15K & 4.8K & 4.8K & 61K \\
        \midrule
        $q^i \rightarrow c^i$ & NIGHTS & Misc. & 16K & 2K & 2K & 40K \\
        \bottomrule
    \end{tabular}
    
\end{table*}

%% file: supp_tables/reasoning_subset.tex
\begin{table*}[t]
    \centering
    \small
    \caption{Summary of the \textbf{Reasoning Subset} in M-BEIR-CoT.}
    \label{tab:reasoning_subset}
    \begin{tabular}{lllrrrr}
        \toprule
        Task & Dataset & Domain & \# Train & \# Dev & \# Test & \# Pool \\
        \midrule
        \multirow{2}{*}{$(q^i,q^t) \rightarrow c^i$} & CIRR & Daily & 24.9K & - & - & 21K \\
        & FashionIQ & Fashion & 15.6K & - & - & 18K \\
        \midrule
        \multirow{2}{*}{$(q^i,q^t) \rightarrow c^t$} & InfoSeek & Wiki & 241K & 3.4K & 6.5K & 1M \\
        & OVEN & Wiki & 294K & 11.7K & 23.3K & 1M \\
        \bottomrule
    \end{tabular}
\end{table*}

%% file: supp_tables/unseen.tex
\begin{table*}[t]
\centering
\caption{\textbf{Summary of the Unseen Dataset.}}
  
\resizebox{.8\textwidth}{!}{  \setlength{\tabcolsep}{1.5mm}{
  \begin{tabular}{lcccc}
    \toprule
    \textbf{Dataset} & \textbf{Image Source} & \textbf{Task} & \textbf{Query Format} & \textbf{Candidate Format}\\
    \midrule
    \multirow{2}{*}{ShareGPT4V} & \multirow{2}{*}{SA-1B} & $q^t \to c^i $ & \texttt{<long text>} & \texttt{<image>} \\
    &  & $q^i \to c^t $ & \texttt{<image>} & \texttt{<long text>} \\
     \midrule
     \multirow{2}{*}{Urban-1K} & \multirow{2}{*}{MSCOCO} & $q^t \to c^i $ & \texttt{<long text>} & \texttt{<image>} \\
    &  & $q^i \to c^t $ & \texttt{<image>} & \texttt{<long text>} \\
     \midrule
     \multirow{2}{*}{Flickr30K} & \multirow{2}{*}{Flickr} & $q^t \to c^i $ & \texttt{<short text>} & \texttt{<image>} \\
    &  & $q^i \to c^t $ & \texttt{<image>} & \texttt{<short text>} \\
    \midrule 
    CIRCO & MSCOCO unlabeled set & $(q^i, q^t) \to c^i $ & \texttt{<image><relative caption>} & \texttt{<image>} \\
    \midrule
    GeneCIS & MSCOCO & $(q^i, q^t) \to c^i $ & \texttt{<image><relative caption>} & \texttt{<image>} \\
    \midrule
    Visual Dialog & MSCOCO & $q^{\text{dialog}} \to c^i $ & \texttt{<Q$_1$><A$_1$>}$\cdots$\texttt{<Q$_\text{j}$><A$_\text{j}$>} & \texttt{<image>} \\
    \midrule 
    Visual Storytelling & Flickr & $ (q^i \oplus q^t) \to c^i $ & \texttt{<text$_1$><image$_1$>}$\cdots$\texttt{<text$_\text{j}$>} & \texttt{<image>} \\
    \midrule
    \multirow{2}{*}{MT-FIQ} & \multirow{2}{*}{FashionIQ} & \multirow{2}{*}{$ (q^i \oplus q^t) \to c^i $} & \multirow{2}{*}{\parbox{7cm}{\texttt{<image$_1$><relative caption$_1$>}$\cdots\\$\texttt{<image$_\text{j}$><relative caption$_\text{j}$>}}} & \multirow{2}{*}{\texttt{<image>}} \\
    & & & & \\
    \midrule
    CC-Neg & CC3M & ITM & \texttt{<image>} & \texttt{<text>} 
    \\
    \midrule
    Sugar-Crepe & MSCOCO & ITM & \texttt{<image>} & \texttt{<text>} \\
    \bottomrule
  \end{tabular}
}
}

\label{tab:unseen_dataset}
\end{table*}

%% file: supp_tables/instructions.tex
\begin{table*}[ht]
\centering
\resizebox{.86\textwidth}{!}{  \setlength{\tabcolsep}{0.6mm}{
  \begin{tabular}{lll}
    \toprule
    \textbf{Task} & \textbf{Dataset} & \textbf{Instruction} \\
    \midrule
    \multirow{12}{*}{$q^t \to c^i$} & \multirow{4}{*}{VisualNews} & Identify the news-related image in line with the described event. \\
     & & Display an image that best captures the following caption from the news.\\
    & & Based on the caption, provide the most fitting image for the news story. \\
    & & I want you to retrieve an image of this news caption. \\
    \cmidrule(r){2-3}
    & \multirow{4}{*}{MSCOCO} & Find me an everyday image that matches the given caption.\\
    & & Identify the image showcasing the described everyday scene. \\
    & & I want to retrieve an image of this daily life description.\\
    & & Show me an image that best captures the following common scene description.\\
     \cmidrule(r){2-3}
     & \multirow{4}{*}{Fashion200K} & Based on the following fashion description, retrieve the best matching image.\\
     & & Match the provided description to the correct fashion item photo.\\
     & & Identify the fashion image that aligns with the described product.\\
     & & You need to identify the image that corresponds to the fashion product description provided.\\
     \midrule
     \multirow{4}{*}{$q^t \to c^t $} & \multirow{4}{*}{WebQA} & Retrieve passages from Wikipedia that provide answers to the following question.\\
     & & You have to find a Wikipedia paragraph that provides the answer to the question.\\
     & & I want to find an answer to the question. Can you find some snippets that provide evidence from Wikipedia?\\
     & & I'm looking for a Wikipedia snippet that answers this question.\\
     \midrule
     \multirow{8}{*}{$q^t \to (c^i, c^t) $} & \multirow{4}{*}{EDIS} & Find a news image that matches the provided caption.\\
     & & Identify the news photo for the given caption.\\
     & & Can you pair this news caption with the right image?\\
     & & I'm looking for an image that aligns with this news caption. \\
     \cmidrule(r){2-3}
      & \multirow{4}{*}{WebQA} & Find a Wikipedia image that answers this question.\\
     & & Provide with me an image from Wikipedia to answer this question. \\
     & & I want to know the answer to this question. Please find the related Wikipedia image for me. \\
     & & You need to retrieve an evidence image from Wikipedia to address this question. \\
     \midrule
     \multirow{12}{*}{$q^i \to c^t $} & \multirow{4}{*}{VisualNews} & Find a caption for the news in the given photo.\\
     & & Based on the shown image, retrieve an appropriate news caption. \\
     & & Provide a news-related caption for the displayed image. \\
     & & I want to know the caption for this news image. \\
     \cmidrule(r){2-3}
     & \multirow{4}{*}{MSCOCO}                  & Find an image caption describing the following everyday image. \\
     & & Retrieve the caption for the displayed day-to-day image. \\
     & & Can you find a caption talking about this daily life image? \\
     & & I want to locate the caption that best describes this everyday scene image. \\
      \cmidrule(lr){2-3}
     & \multirow{4}{*}{Fashion200K} & Find a product description for the fashion item in the image.  \\
     & & Based on the displayed image, retrieve the corresponding fashion description.  \\
     & & Can you retrieve the description for the fashion item in the image?  \\
     & & I want to find a matching description for the fashion item in this image.  \\
     \midrule
     \multirow{4}{*}{$q^i \to c^i $} & \multirow{4}{*}{NIGHTS} & Find a day-to-day image that looks similar to the provided image.\\
     & & Which everyday image is the most similar to the reference image? \\
     & & Find a daily life image that is identical to the given one. \\
     & & You need to identify the common scene image that aligns most with this reference image. \\
     \midrule 
     \multirow{8}{*}{$(q^i, q^t) \to c^t $} & \multirow{4}{*}{OVEN} & Retrieve a Wikipedia paragraph that provides an answer to the given query about the image.\\
     & & Determine the Wikipedia snippet that identifies the visual entity in the image. \\
     & & I want to find a paragraph from Wikipedia that answers my question about this image. \\
     & & You have to find a Wikipedia segment that identifies this image's subject. \\
     \cmidrule(lr){2-3}
     & \multirow{4}{*}{InfoSeek} & Retrieve a Wikipedia paragraph that provides an answer to the given query about the image.\\
     & & Determine the Wikipedia snippet that matches the question of this image. \\
     & & I want to find a paragraph from Wikipedia that answers my question about this image. \\
     & & You have to find a Wikipedia segment that answers the question about the displayed image. \\
     \midrule 
     \multirow{8}{*}{$(q^i, q^t) \to c^i $} & \multirow{4}{*}{FashionIQ}  & Find a fashion image that aligns with the reference image and style note. \\
     & & With the reference image and modification instructions, find the described fashion look. \\
     & & Given the reference image and design hint, identify the matching fashion image. \\
     & & I'm looking for a similar fashion product image with the described style changes. \\
     \cmidrule(lr){2-3}
     & \multirow{4}{*}{CIRR}  & Retrieve a day-to-day image that aligns with the modification instructions of the provided image. \\
     & & Pull up a common scene image like this one, but with the modifications I asked for. \\
     & & Can you help me find a daily image that meets the modification from the given image? \\
     & & I'm looking for a similar everyday image with the described changes. \\
    \midrule
    \multirow{8}{*}{$(q^i, q^t) \to (c^i, c^t) $} & \multirow{4}{*}{OVEN} & Retrieve a Wikipedia image-description pair that provides evidence for the question of this image. \\
    & & Determine the Wikipedia image-snippet pair that clarifies the entity in this picture. \\
    & & I want to find an image and subject description from Wikipedia that answers my question about this image. \\
    & & I want to know the subject in the photo. Can you provide the relevant Wikipedia section and image? \\
    \cmidrule(lr){2-3}
    & \multirow{4}{*}{InfoSeek} & Retrieve a Wikipedia image-description pair that provides evidence for the question of this image. \\
    & & Determine the Wikipedia image-snippet pair that matches my question about this image. \\
    & & I want to find an image and subject description from Wikipedia that answers my question about this image. \\
    & & I want to address the query about this picture. Please pull up a relevant Wikipedia section and image. \\
    \bottomrule
  \end{tabular}
}
}
\caption{\textbf{Summary of the M-BEIR instructions.}
}
\vspace{-2em}
\label{tab:instruction}
\end{table*}